\documentclass[sigconf]{aamas}

\usepackage{subcaption}
\usepackage{newfloat}
\usepackage{tabularx}
\usepackage{graphicx}

\usepackage{balance} 

\usepackage{listings}
\DeclareCaptionStyle{ruled}{labelfont=normalfont,labelsep=colon,strut=off}
\floatstyle{ruled}
\newfloat{listing}{tb}{lst}{}
\floatname{listing}{Listing}

\usepackage{amsmath}
\usepackage{booktabs}
\usepackage{float}  

\usepackage{graphicx}  
\usepackage{colortbl}

\makeatletter
\gdef\@copyrightpermission{
  \begin{minipage}{0.2\columnwidth}
   \href{https://creativecommons.org/licenses/by/4.0/}{\includegraphics[width=0.90\textwidth]{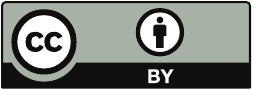}}
  \end{minipage}\hfill
  \begin{minipage}{0.8\columnwidth}
   \href{https://creativecommons.org/licenses/by/4.0/}{This work is licensed under a Creative Commons Attribution International 4.0 License.}
  \end{minipage}
  \vspace{5pt}
}
\makeatother

\setcopyright{ifaamas}
\acmConference[AAMAS '25]{Proc.\@ of the 24th International Conference
on Autonomous Agents and Multiagent Systems (AAMAS 2025)}{May 19 -- 23, 2025}
{Detroit, Michigan, USA}{Y.~Vorobeychik, S.~Das, A.~Nowé  (eds.)}
\copyrightyear{2025}
\acmYear{2025}
\acmDOI{}
\acmPrice{}
\acmISBN{}

\acmSubmissionID{<<OpenReview submission id>>}

\title[AAMAS-2025 Formatting Instructions]{An Extended Benchmarking of Multi-Agent Reinforcement Learning Algorithms in Complex Fully Cooperative Tasks}

\author{George Papadopoulos}
\authornote{Equal Contribution.}
\affiliation{
  \institution{University of Piraeus}
  \city{Piraeus}
  \country{Greece}
  }
\email{georgepap@unipi.gr}

\author{Andreas Kontogiannis}
\authornotemark[1]
\affiliation{
  \institution{NTUA \& Archimedes AI}
  \city{Athens}
  \country{Greece}
  }
  \email{andreaskontogiannis@mail.ntua.gr}

\author{Foteini Papadopoulou}
\affiliation{
  \institution{Radboud University}
  \city{Nijmegen}
  \country{Netherlands}
  }
\email{foteini.papadopoulou@ru.nl}

\author{Chaido Poulianou}
\affiliation{
  \institution{University of Piraeus}
  \city{Piraeus}
  \country{Greece}
  }
  \email{cpoulianou@uth.gr}

\author{Ioannis Koumentis}
\affiliation{
  \institution{University of Piraeus}
  \city{Piraeus}
  \country{Greece}
  }
  \email{iokoumen@unipi.gr}

\author{George Vouros}
\affiliation{
  \institution{University of Piraeus}
  \city{Piraeus}
  \country{Greece}
  }
  \email{georgev@unipi.gr}

\begin{abstract}
Multi-Agent Reinforcement Learning (MARL) has recently emerged as a significant area of research. However, MARL evaluation often lacks systematic diversity, hindering a comprehensive understanding of algorithms' capabilities. In particular, cooperative MARL algorithms are predominantly evaluated on benchmarks such as SMAC and GRF, which primarily feature team game scenarios without assessing adequately various aspects of agents' capabilities required in fully cooperative real-world tasks such as multi-robot cooperation and warehouse, resource management, search and rescue, and human-AI cooperation. Moreover, MARL algorithms are mainly evaluated on low dimensional state spaces, and thus their performance on high-dimensional (e.g., image) observations is not well-studied. To fill this gap, this paper highlights the crucial need for expanding systematic evaluation across a wider array of existing benchmarks. To this end, we conduct extensive evaluation and comparisons of well-known MARL algorithms on complex \textit{fully} cooperative benchmarks, including tasks with \textit{images} as agents' observations. Interestingly, our analysis shows that many algorithms, hailed as state-of-the-art on SMAC and GRF, may underperform standard MARL baselines on fully cooperative benchmarks. Finally, towards more systematic and better evaluation of cooperative MARL algorithms, we have open-sourced PyMARLzoo+, an extension of the widely used (E)PyMARL libraries, which addresses an open challenge from~\cite{pettingzoo}, facilitating seamless integration and support with all benchmarks of PettingZoo, as well as Overcooked, PressurePlate, Capture Target and Box Pushing.
\end{abstract}


\newcommand{\BibTeX}{\rm B\kern-.05em{\sc i\kern-.025em b}\kern-.08em\TeX}

\begin{document}


\pagestyle{fancy}
\fancyhead{}


\maketitle 


\section{Introduction}

In \textit{fully} cooperative Multi-Agent Reinforcement Learning (MARL) problems, the goal is to train learnable agents in order to maximize their shared cumulative reward, {through excessive coordination, sharing of tasks, collaborative exploration, with appropriate decisions for timing  and action, and sharing of capabilities}. 
Fully cooperative MARL is of remarkable interest, as it can naturally model many real-world applications, including multi-robot collaboration~\cite{burgard2000collaborative} and warehouse~\cite{papoudakis2021benchmarking}, search and rescue~\cite{rahman2022adversar}, human-AI coordination~\cite{overcooked}, air traffic management \cite{kontogiannis2023inherently}, logistics networks \cite{logistics}, and supply-chain optimization \cite{supply_chain}. Recently, cooperative MARL algorithms {are mainly evaluated in settings} where adversarial non-learnable agents  that interact with the learnable cooperative ones exist: This has received a surge of approaches and methodologies, e.g., see \cite{liu2021coach, yang2022transformer, jeon2022maser, sun2024decision}, also drawing motivation from multiplayer video-games and team sports.  

Despite recent efforts~\cite{papoudakis2021benchmarking, yu2020benchmarking, bettini2024benchmarl, hu2022marllib} aiming to provide a comprehensive understanding of standard cooperative MARL algorithms' capabilities through benchmarking, MARL evaluation still lacks systematic diversity and reliability for the following reasons: 
\begin{itemize}
    \item Most state-of-the-art (SoTA) cooperative MARL algorithms are predominantly evaluated, and possibly overfit, as pointed out in~\cite{protocol}, on specific cooperative-competitive benchmarks where a team of cooperative learning agents competes against a team of bots with fixed policies, namely SMAC~\cite{smac, smac2} and GRF~\cite{grf}. However, we argue that these benchmarks {do not allow adequate evaluation of subtle issues involved in} \textit{fully} cooperative MARL, including: {excessive coordination and exploration capabilities, sharing of capabilities, appropriate timing in the execution of actions, complimentary observability, scaling to large numbers of agents, and possibly with sparse rewards. For instance, tasks from the LBF benchmark~\cite{papoudakis2021benchmarking},and particularly those with large grids, effectively capture more diverse requirements of fully cooperative multi-robot collaboration: They require agents to first engage in \textit{extensive joint exploration} to identify a \textit{certain} food target, followed by coordinated joint actions where all agents must \textit{simultaneously} consume that target. We conjecture that these aspects are crucial for fully cooperative, real-world tasks.} On the other hand, these benchmarks emphasize developing skills {that do not play dominant roles} for fully cooperative tasks, such as countering the opponent team (e.g., surviving enemy attacks in SMAC or tackling opponents and preventing goals in GRF). 
    \item Most MARL algorithms are evaluated solely on tasks with {low-dimensional} (mostly tabular) state spaces, and thus their effectiveness on real-world, high-dimensional, image-based observations has not been studied.
    \item  MARL evaluation  does not often report the training times of the proposed algorithms, so the results cannot be interpreted as a function of the compute budget used~\cite{protocol}.
\end{itemize}

To address the above challenges, the main contributions of this paper are the following: 
\textbf{(1)} Our paper investigates the effectiveness of established MARL algorithms and contributes a comprehensive, \textbf{updated empirical evaluation and comparison} of MARL algorithms, including algorithms that have demonstrated SoTA performance in SMAC and GRF, across a wide array of complex \textbf{fully cooperative} benchmarks.
\textbf{(2)} To our knowledge, our work is the first to include tasks with \textit{images} representing \textit{high-dimensional observations} in MARL benchmarking. 
\textbf{(3)} We contribute an open-source Python MARL framework, namely \textit{PyMARLzoo+} \footnote{The source code is available at: {\color{blue}\url{https://github.com/AILabDsUnipi/pymarlzooplus}}}, which extends the (E)PyMARL frameworks \cite{smac,papoudakis2021benchmarking} (widely used in developing established MARL algorithms, such as \cite{qmix,qplex,wang2020roma,cds,jeon2022maser}), facilitating seamless integration with all PettingZoo tasks, thus addressing an open challenge from \cite{pettingzoo} (EPyMARL supports only the MPE PettingZoo tasks, which were already integrated in \cite{papoudakis2021benchmarking}). In addition to the already integrated MPE \cite{mpe,mpe2}, LBF \cite{lbf} and RWARE \cite{papoudakis2021benchmarking} benchmarks, our PyMARLzoo+ also integrates the complex fully cooperative Overcooked \cite{overcooked}, PressurePlate \cite{ai_albrecht_edinburgh, gupta2023cammarl}, Capture Target \cite{omidshafiei2017deep, xiao2020macro} and BoxPushing \cite{xiao2020learning, xiao2022asynchronous} benchmarks.
\textbf{(4)} Our work provides benchmarking of tasks that are indicative of a wide array of real-world applications, and of the evaluation of diverse requirements for fully cooperative MARL, in terms of joint exploration and coordination. 
\textbf{(5)} Our experimental findings demonstrate that many algorithms, which have been SoTA in SMAC and GRF, may underperform standard MARL algorithms in fully cooperative MARL tasks; thus validating the possible overfitting issues of the current MARL evaluation highlighted in \cite{protocol}. 
Furthermore, we point out fully cooperative tasks that are very hard to solve with the existing SoTA methods.   
\textbf{(6)} Our benchmarking is the first to report the training times of algorithms as a further measure of the algorithm's performance, so that the reported results are also interpreted as a function of the compute budget used.

\section{Preliminaries}

\subsection{Dec-POMDPs}
Fully cooperative MARL is formulated as a Dec-POMDP \cite{oliehoek2016concise}.
A Dec-POMDP for an $N$-agent task is a tuple $\langle S, A, P, r, F, O, N, \gamma \rangle$, where $S$ is the state space, $A$ is the joint action space $A=A_1 \times \dots \times A_N$, $P(s' \mid s, a) : S \times A \rightarrow [0,1]$ is the state transition function, $r(s, a): S \times A \rightarrow {\mathbb{R}}$ is the \textit{shared} reward function and $\gamma \in [0,1)$ is the discount factor. Assuming partial observability, each agent at time step $t$ does not have access to the full state, yet it samples observations $o_t^i \in O_i$ according to the observation function $F_i(s): S \rightarrow O_i$. The joint observations are denoted by $o \in O$ and are sampled according to $F=\prod_iF_i$.
The action-observation history for agent $i$ at time $t$ is denoted by $h_t^i \in H_i$, which includes action-observation pairs until $t$-$1$ and $o^i_t$, on which the agent can condition its individual stochastic policy $\pi_{\theta_i}^i(a_t^i \mid h_t^i): H_i \times A_i \rightarrow [0,1]$,  parameterised by $\theta_{i}$. 
The joint policy is denoted by $\pi_{\theta}$, with parameters  $\theta \in \Theta$ . 
The objective is to find an optimal joint policy which satisfies the optimal value function $V^*({s}) = \max_{\theta} \mathbb{E}_{a \sim \pi_{\theta}, s' \sim P(\cdot|s,a), o \sim F(s)} \left [\sum_{t=0}^{\infty} \gamma^t r_{t} \right ]$.

\subsection{Assumptions of interest}

In addition to \textit{partial observability}, in our benchmark analysis we consider and study the following assumptions of interest: (a) we utilize the celebrated \textit{CTDE} MARL schema \cite{mpe2}, which has been widely adopted by the cooperative MARL community \cite{protocol, marl_book} as it enables conditioning approximate value functions on privileged information in
a computationally tractable manner, (b) we evaluate \textit{fully cooperative tasks}, that is, tasks where there are no adversarial non-learnable agents (e.g., bots), or teams of opponent agents, interacting with the learnable agents, (c) we also evaluate complex tasks with \textit{sparse-reward settings} that require excessive joint exploration and cooperation to be solved, (d) we also study tasks with \textit{image-based}, \textit{high-dimensional state and observation spaces} (as in real-world tasks), which have been very frequent in single-agent RL but not in MARL evaluation, and (e) as in \cite{yu2020benchmarking, papoudakis2021benchmarking} we assume that agents
\textit{lack explicit communication channels during execution}.    

{
\small
\begin{table*}[t]
\centering
\begin{tabularx}{\textwidth}{|>{\centering\arraybackslash}X|>{\centering\arraybackslash}X|>{\centering\arraybackslash}X|>{\centering\arraybackslash}X|>{\centering\arraybackslash}X|>{\centering\arraybackslash}X|>{\centering\arraybackslash}X|}
\hline
\textbf{Algorithm} & \textbf{Evaluated in} & \textbf{On/Off-Policy} & \textbf{RL Optimization} & \textbf{Network Architectures} & \textbf{Intrinsic Exploration}  \\
\hline
\textbf{QMIX} \cite{qmix} & SMAC \cite{smac}, GRF \cite{grf}, MPE \cite{mpe}, LBF \cite{lbf}, RWARE \cite{papoudakis2021benchmarking}
& Off-policy
& Value-based
& RNN \& MLP
& No
\\
\hline
\textbf{MAA2C} \cite{papoudakis2021benchmarking} & SMAC \cite{smac}, GRF \cite{grf}, MPE \cite{mpe}, LBF \cite{lbf}, RWARE \cite{papoudakis2021benchmarking}
& On-policy
& Actor-Critic
& RNN \& MLP
& No
\\
\hline
\textbf{COMA} \cite{coma} & SMAC \cite{smac}, MPE \cite{mpe}, LBF \cite{lbf}, RWARE \cite{papoudakis2021benchmarking}
& On-policy
& Policy Gradient
& RNN \& MLP
& No
\\
\hline
\textbf{MAPPO} \cite{mappo} & SMAC \cite{smac}, GRF \cite{grf}, MPE \cite{mpe}, LBF \cite{lbf}, RWARE \cite{papoudakis2021benchmarking}
& On-policy
& Actor-Critic
& RNN \& MLP
& No
\\
\hline
\textbf{QPLEX} \cite{qplex} & SMAC \cite{smac}
& Off-policy
& Value-based
& RNN \& MLP
& No
\\
\hline
\textbf{HAPPO} \cite{happo} & SMAC \cite{smac}, MA MuJoCo \cite{facmac}
& On-policy
& Actor-Critic
& RNN \& MLP
& No
\\
\hline
\textbf{MAT-DEC} \cite{mat} & SMAC \cite{smac}, GRF \cite{grf}, MA MuJoCo \cite{facmac}
& On-policy
& Actor-Critic
& Transformer \& MLP
& No
\\
\hline
\textbf{EMC} \cite{emc} & SMAC \cite{smac}
& Off-policy
& On top of QPLEX
& RNN \& MLP
& \textit{Yes}: curiosity-driven
\\
\hline
\textbf{MASER} \cite{maser} & SMAC \cite{smac}
& Off-policy
& On top of QMIX
& RNN \& MLP
& \textit{Yes}: subgoal generation
\\
\hline
\textbf{EOI} \cite{eoi} & GRF \cite{grf}, MAgent \cite{zheng2018magent}
& Off-policy
& On top of MAA2C
& RNN \& MLP
& \textit{Yes}: individuality
\\
\hline
\textbf{CDS} \cite{cds} & SMAC \cite{smac}, GRF \cite{grf}
& Off-policy
& On top of QPLEX
& RNN \& MLP
& \textit{Yes}: diversity \& information sharing
\\
\hline
\end{tabularx}
\caption{Summary of the selected MARL algorithms}
\label{table:algorithms}
\end{table*}
}

\section{Algorithms}\label{algorithms}

In our benchmark analysis, we evaluate and compare a wide range of well-known MARL algorithms (see Table \ref{table:algorithms}). The selection of these algorithms is based on the following important (but not all-encompassing) criteria: (1) they have been widely used as competitive baselines by the MARL community in recent works, (2) they have achieved SoTA performance in cooperative-competitive MARL benchmarks, such as SMAC and GRF, and in our analysis we aim to test their performance in fully cooperative tasks, (3) they are based on improving joint exploration in sparse-reward tasks, and (4) they present diversity in the RL optimization part (e.g., being value-based or actor-critic based, utilizing standard recurrent architectures or transformer-based).    

\section{Fully Cooperative MARL Benchmarks}\label{benchmarks}
In this section, we highlight complex, partial observable, fully cooperative tasks from eight MARL benchmarks that we believe to be of significant interest for MARL research.
These tasks represent a wide range of real-world applications and test diverse requirements for fully cooperative MARL (see also Table \ref{table:benchmarks}).

{
\small
\begin{table*}[t]
\centering
\begin{tabularx}{\textwidth}{|>{\centering\arraybackslash}X|>{\centering\arraybackslash}X|>{\centering\arraybackslash}X|}
\hline
\cellcolor{blue!25}\textbf{MARL Benchmark} & \cellcolor{red!25}\textbf{Interesting Challenges} & \cellcolor{orange!25}\textbf{Insights for Real-world Applications} \\
\hline
\textit{Entombed Cooperative} (PettingZoo) & 
- RGB observations
 & 
- exploration in space missions
\\
& 
- dead-ends
 & 
- search \& rescue
\\
& 
- excessive coordination to go to opposite sides in order to break through walls symmetrically
& 
- collective wildfire movement
    \\
\hline
\textit{Pistonball} (PettingZoo) & 
- RGB observations
& 
- assembly line coordination
 \\
& 
- excessive coordination for emergent behavior and synchronization
& 
- dance and performance choreography
 \\
\hline

\textit{Cooperative Pong} (PettingZoo) & 
- RGB observations
& 
- collaborative video games
\\
& 
- excessive coordination due to asymmetry 
& 
- performing arts
\\
& between the agents
& - sports collaboration
\\

\hline
Overcooked (\textit{Cramped Room},  & 
- sparse reward
& 
- kitchen automation
\\
\textit{Asymmetric Advantages}, \textit{Coordination Ring})& 
- delivering tasks as quickly as possible
& 
- urgent multi-robot tasks in confined spaces
\\
& 
- agents operating in a confined space
& 
\\
& 
- agent collision avoidance
& 
\\
& 
- splitting tasks on the fly
& 
\\
\hline
{Pressure Plate}  & 
- sparse reward 
& 
- multi-robot collaboration 
\\
& 
- excessive exploration and coordination to 
& 
- escape-rooms-like tasks 
\\
& 
sequentially unlock each room
& 
- team-based problem solving
\\
\hline
\textit{Spread} (MPE) & 
- agent collision avoidance 

- optimal landmark coverage

- very complex if $N$ is large
& 
- traffic management 

- distributed sensor networks 

- urban planning 
\\
\hline
{LBF}  & 
- sparse reward 

- exploration to identify the food target

- excessive coordination to eat the target simultaneously
& 
 - multi-robot collaboration 

- resource management in supply chains 

- disaster response coordination 
\\
\hline
{RWARE}  & 
- sparse reward and high-dimensional observations 

- excessive coordination to execute a specific sequence of actions, at the right time, without immediate feedback 
& 
- robot warehousing 

- logistics management 
\\
\hline
\end{tabularx}
\caption{Summary of the fully cooperative, partially observable, benchmark tasks}
\label{table:benchmarks}
\end{table*}
}

\subsection{PettingZoo}

PettingZoo \cite{pettingzoo} is a Python library consisting of several MARL tasks. 
In our analysis, we utilize the following fully cooperative PettingZoo benchmarks: \textit{Entombed Cooperative}, along with \textit{Pistonball} and \textit{Cooperative Pong}. These tasks allow challenging scenarios that provide useful testbeds for fully cooperative 
MARL, using high-dimensional, RGB-image-based observation spaces. Indicative references to the PettingZoo benchmark include \citep{pettingzoo,terry2020revisiting,terry2020multiplayer,siu2021dynamic,chen2024generative}.

\subsubsection{Entombed Cooperative}

Here, agents must extensively coordinate to progress as far as possible into a procedurally generated maze. Each agent needs to quickly navigate down a constantly evolving maze, where only part of the environment is visible. If an agent becomes trapped, they lose. Agents can easily find themselves in dead-ends, only escapable through rare power-ups. 
A major challenge is that optimal coordination requires agents to position themselves on opposite sides of the map, as power-ups appear on one side or the other but can be used to break through walls symmetrically.

\subsubsection{Pistonball}

Pistonball is a physics-based fully cooperative game in which the goal is to move a ball to the left boundary of the game area by controlling a set of vertically moving pistons. 
The main challenge lies in achieving highly coordinated, emergent behavior to optimize performance in the environment. Pistonball uses a realistic physics engine, comparable to the game Angry Birds, adding further complexity to the required agent coordination.

\subsubsection{Cooperative Pong}

Cooperative Pong is a fully cooperative variant of the classic Pong game where two agents control paddles on opposite sides of the screen, aiming to keep the ball in play. The challenge lies in the asymmetry between the agents—particularly the right paddle, which has a more complex tiered shape—and their limited observation space, restricted to their own half of the screen. This setup requires agents to coordinate excessively and develop cooperative strategies to maximize play time.

\subsection{Overcooked}

Overcooked \cite{overcooked} is a fully cooperative MARL benchmark, which has been developed to address human-AI coordination. The objective is to deliver soups as quickly as possible, with agents required to place up to three ingredients in a pot, wait for the soup to cook, and then deliver it. The tight space introduces significant challenges in terms of coordination, as agents must split tasks on the fly, avoid collisions, and coordinate effectively in order to achieve high reward. In addition, the rewards are sparse, making the environment even more challenging. Indicative references that use this benchmark include \cite{pettingzoo,siu2021dynamic,chen2024generative}. We utilize the following three different layouts.

\subsubsection{Cramped Room}

Agents operate in a cramped room, that is, a small room without obstacles. This layout focuses on the agents' ability to maneuver in a limited space, requiring precise navigation to avoid physical collisions with the other agent.

\subsubsection{Asymmetric Advantages}

Agents operate in an asymmetric room, where each agent works in its own distinct space. In this layout, the challenge lies in recognizing and exploiting the differing strengths of each agent, such as speed or access to certain kitchen resources. 
This task tests the agents' ability to adapt their strategies to complement each other's capabilities, ensuring that each player's strengths are used effectively to enhance overall team performance.

\subsubsection{Coordination Ring}

Agents operate in a room with a ring. They must coordinate their actions to collect onions from the bottom left corner of the room, make soups in the center left, and deliver the dishes to the top right corner of the ring. This task is the most challenging among the three.

\subsection{PressurePlate}

PressurePlate \cite{ai_albrecht_edinburgh} is a fully cooperative environment, with sparse-reward settings, set within a 2D grid-world composed of multiple locked rooms that can be unlocked when an agent stands on the corresponding pressure plate, culminating in a final room containing a goal chest. The primary challenge for agents lies in effectively coordinating their movements and positions to sequentially unlock each room and ultimately reach the chest.  
Indicative references that use this benchmark task include \cite{ai_albrecht_edinburgh, gupta2023cammarl}.

\subsection{Multi-agent Particle Environment (MPE)}

In our analysis, we utilize the well-known \textit{Spread} task of the MPE benchmark \cite{mpe2} that focus on effective navigation of particle agents. Indicative references that use this benchmark task include \cite{papoudakis2021benchmarking,ruan2022gcs_yalidu,zhong2024heterogeneous,christianos2021scaling}. In this task,
$N$ agents must cover $N$ landmarks while avoiding collisions. Agents are rewarded for staying close to landmarks and penalized for collisions, creating a trade-off between coordination and collision avoidance. The task becomes more complex as $N$ increases. Compared to \cite{papoudakis2021benchmarking}, the evaluated tasks are more challenging by using more agents and fewer training steps.


\subsection{Level Based Foraging (LBF)}

Level-Based Foraging (LBF) \cite{lbf} features fully cooperative grid-world environments where agents, assigned levels, must move and collect food by coordinating their actions. Agents can collect food only if their combined levels meet or exceed the food's level. The main challenge is the sparse rewards, requiring agents to coordinate closely to collect food simultaneously. Unlike previous work \cite{papoudakis2021benchmarking}, we evaluate more complex LBF tasks, even with a larger number of agents, requiring extensive exploration and coordination. Indicative references that use this benchmark include \cite{overcooked,yang2024hierarchical,fosong2024learning,hong2024learning}.

{
\small
\begin{table}[t]
\centering
\begin{tabular}{|>{\centering\arraybackslash}p{0.3\linewidth}|>{\centering\arraybackslash}p{0.3\linewidth}|}
\hline
\textbf{Benchmark} & \textbf{Number of Newly Integrated Tasks} \\
\hline
PettingZoo & 49 \\
Overcooked & 5 \\
Pressure Plate & 3 \\ 
Capture Target & 1 \\
Box Pushing & 1 \\
\hline
\end{tabular}
\caption{Newly integrated tasks (in addition to MPE, LBF and RWARE) in PyMARLzoo+.}
\label{table: envs_all}
\end{table}
}

\subsection{Multi-Robot Warehouse (RWARE)}

The Multi-Robot Warehouse (RWARE) environment simulates a fully cooperative, partially observable grid-world where agents must find and deliver shelves to workstations. With limited sight and partial observations, agents face challenges such as sparse rewards, only given after successful shelf deliveries. This requires precise action sequences, effective exploration, and excessive agent coordination. Unlike previous work \cite{papoudakis2021benchmarking}, this study evaluates more challenging RWARE tasks in hard mode, demanding efficient exploration and excessive cooperation among agents. Indicative references that use RWARE include \cite{pettingzoo,terry2020revisiting,terry2020multiplayer}.

\subsection{Capture Target}

Capture Target  \cite{omidshafiei2017deep, xiao2020macro}, a fully cooperative multi-agent single-target task, presents significant challenges, aiming multiple agents to locate and capture a flickering target on a grid. Excessive coordination is essential, as the target is only captured when all agents converge on its location simultaneously, despite limited visibility.

\subsection{Box Pushing}

Box Pushing \cite{xiao2020learning, xiao2022asynchronous} is a fully cooperative grid environment where two agents must collaborate to move three boxes—two small and one large—to a target area. The main challenge is that the large box, yielding the highest reward, can only be moved if both agents coordinate by positioning themselves in parallel cells and pushing simultaneously, making precise timing and teamwork essential. 

{
\tiny
\begin{table*}[t]

\centering
\begin{tabular}{@{} lccccccccccc @{}}

\toprule

\textbf{Tasks\textbackslash Algorithms} & \textbf{QMIX} & \textbf{QPLEX} & \cellcolor{green!25}\textbf{MAA2C} & \textbf{MAPPO} & \textbf{HAPPO} & \textbf{MAT-DEC} & \textbf{COMA} & \textbf{EOI} & \textbf{MASER} & \textbf{EMC} & \textbf{CDS} \\

\midrule

\textbf{2s-8x8-3p-2f} & \(0.94 \pm 0.09\) & \(0.63 \pm 0.48\) & \(\textbf{0.98} \pm \textbf{0.02}\) & \(0.64 \pm 0.34\) & \(0.00 \pm 0.00\) & \(0.24 \pm 0.22\) & \(0.00 \pm 0.00\) & \(0.00 \pm 0.00\) & \(0.00 \pm 0.00\) & \(0.00 \pm 0.00\) & \(0.00 \pm 0.00\) \\

\textbf{2s-9x9-3p-2f} & \(0.00 \pm 0.00\) & \(\textbf{0.60} \pm \textbf{0.49}\) & \(\textbf{0.59} \pm \textbf{0.38}\) & \(0.18 \pm 0.35\) & \(0.00 \pm 0.00\) & \(0.34 \pm 0.24\) & \(0.00 \pm 0.00\) & \(0.00 \pm 0.00\) & \(0.00 \pm 0.00\) & \(0.50 \pm 0.50\) & \(0.00 \pm 0.00\) \\

\textbf{2s-12x12-2p-2f} & \(0.89 \pm 0.01\) & \(\textbf{0.98} \pm \textbf{0.00}\) & \(0.81 \pm 0.03\) & \(0.77 \pm 0.04\) & \(0.52 \pm 0.26\) & \(0.55 \pm 0.06\) & \(0.03 \pm 0.02\) & \(0.34 \pm 0.23\) & \(0.01 \pm 0.01\) & \(0.87 \pm 0.03\) & \(0.91 \pm 0.02\) \\

\textbf{4s-11x11-3p-2f} & \(\textbf{0.08} \pm \textbf{0.19}\) & \(0.00 \pm 0.00\) & \(0.00 \pm 0.00\) & \(0.00 \pm 0.00\) & \(0.00 \pm 0.00\) & \(0.00 \pm 0.00\) & \(0.00 \pm 0.00\) & \(0.00 \pm 0.00\) & \(0.00 \pm 0.00\) & \(0.00 \pm 0.00\) & \(0.00 \pm 0.00\) \\

\textbf{7s-20x20-5p-3f} & \(0.01 \pm 0.00\) & \(0.00 \pm 0.00\) & \(\textbf{0.78} \pm \textbf{0.02}\) & \(0.57 \pm 0.18\) & \(0.00 \pm 0.00\) & \(0.29 \pm 0.09\) & \(0.03 \pm 0.01\) & \(0.03 \pm 0.01\) & \(0.01 \pm 0.00\) & \(0.01 \pm 0.00\) & \(0.00 \pm 0.00\) \\

\textbf{8s-25x25-8p-5f} & \(0.02 \pm 0.01\) & \(0.01 \pm 0.00\) & \(\textbf{0.52} \pm \textbf{0.24}\) & \(\textbf{0.41} \pm \textbf{0.23}\) & \(0.00 \pm 0.00\) & \(0.03 \pm 0.03\) & \(0.03 \pm 0.00\) & \(0.07 \pm 0.00\) & \(0.01 \pm 0.00\) & \(0.00 \pm 0.00\) & \(0.00 \pm 0.00\) \\

\textbf{7s-30x30-7p-4f} & \(0.06 \pm 0.05\) & \(0.00 \pm 0.00\) & \(\textbf{0.71} \pm \textbf{0.02}\) & \(0.57 \pm 0.03\) & \(0.00 \pm 0.00\) & \(0.08 \pm 0.06\) & \(0.02 \pm 0.00\) & \(0.04 \pm 0.00\) & \(0.01 \pm 0.00\) & \(0.00 \pm 0.00\) & \(0.00 \pm 0.00\) \\

\bottomrule
\end{tabular}
\caption{Results in {LBF} tasks.}
\label{table:LBF-results}
\end{table*}
}

{
\tiny
\begin{table*}[t]
\centering
\begin{tabular}{@{} lccccccccccc @{}}

\toprule

\textbf{Tasks\textbackslash Algorithms} & \textbf{QMIX} & \textbf{QPLEX} & \textbf{MAA2C} & \textbf{MAPPO} & \cellcolor{green!25}\textbf{HAPPO} & \cellcolor{green!25}\textbf{MAT-DEC} & \textbf{COMA} & \textbf{EOI} & \textbf{MASER} & \textbf{EMC} & \cellcolor{green!25}\textbf{CDS} \\

\midrule

\textbf{tiny-2ag-hard} & \(0.00 \pm 0.00\) & \(0.61 \pm 0.45\) & \(2.25 \pm 0.62\) & \(2.91 \pm 0.82\) & \(1.46 \pm 2.06\) & \(\textbf{6.00} \pm \textbf{8.49}\) & \(0.02 \pm 0.00\) & \(\textbf{6.58} \pm \textbf{3.92}\) & \(0.00 \pm 0.00\) & \(1.66 \pm 0.30\) & \(4.00 \pm 2.30\) \\

\textbf{tiny-4ag-hard} & \(0.00 \pm 0.00\) & \(11.73 \pm 10.81\) & \(6.87 \pm 7.12\) & \(17.1 \pm 5.31\) & \(24.07 \pm 0.71\) & \(\textbf{27.85} \pm \textbf{19.71}\) & \(0.02 \pm 0.00\) & \(12.09 \pm 4.59\) & \(0.00 \pm 0.00\) & \(0.00 \pm 0.00\) & \(\textbf{27.37} \pm \textbf{6.88}\) \\

\textbf{small-4ag-hard} & \(0.01 \pm 0.01\) & \(0.94 \pm 0.63\) & \(1.38 \pm 1.26\) & \(4.14 \pm 2.12\) & \(\textbf{9.69} \pm \textbf{3.19}\) & \(0.00 \pm 0.00\) & \(0.03 \pm 0.00\) & \(0.17 \pm 0.01\) & \(0.02 \pm 0.00\) & \(0.05 \pm 0.00\) & \(4.11 \pm 0.32\) \\

\bottomrule

\end{tabular}
\caption{Results in RWARE tasks.}
\label{tab:RWARE-results}
\end{table*}
}

\section{PyMARLzoo+: PettingZoo, Overcooked, Pressure Plate, Capture Target and Box Pushing now compatible with (E)PyMARL}

To facilitate a comprehensive understanding of MARL algorithms through benchmarking, we open-source \textit{PyMARLzoo+}, a Python framework which extends the widely used (E)PyMARL ~\cite{smac,papoudakis2021benchmarking}, providing an integration of many SoTA algorithms on a plethora of existing MARL benchmarks under a common framework. 
The key features of our framework are presented below:

\paragraph{Newly Integrated Benchmarks.}
Our PyMARLzoo+ integrates 8 MARL benchmarks described in Section \ref{benchmarks}. More specifically, as we illustrate in Table \ref{table: envs_all}, in addition to the MPE, LBF and RWARE benchmarks (already integrated in EPyMARL \cite{papoudakis2021benchmarking}), our framework fully supports all tasks from PettingZoo, along with tasks from Overcooked, Pressure Plate, Capture Target, and Box Pushing.

\paragraph{Newly Integrated Algorithms.}
Our PyMARLzoo+ integrates all the algorithms described in Section \ref{algorithms}. We note that, except for the standard baselines (that is, MAA2C, MAPPO and QMIX) already integrated in EPyMARL, all remaining algorithms were integrated as part of our work. Furthermore, the widely used \textit{Prioritized Replay Buffer} \cite{schaul2015prioritized} has been integrated into the off-policy algorithms.

\paragraph{Image (Observation) encoding.}

We incorporate three pre-trained architectures as image encoders to transform image-based observations into tabular format for policy learning. Specifically, we integrate the following options: \textit{ResNet18}~\cite{he2016deep}, \textit{CLIP}~\cite{radford2021learning}, and \textit{SlimSAM}~\cite{slimsam}.ResNet18, commonly used in single-agent RL (e.g., in \cite{shah2021rrl}), can capture spatial hierarchies, which are beneficial for extracting relevant features from complex visual data. CLIP, also used in single agent RL (e.g., in \cite{clip_in_rl}), utilizes natural language supervision to produce robust and semantically meaningful representations, facilitating model generalization across diverse tasks by linking visual inputs with textual descriptions. Similarly, SlimSAM incorporates a streamlined self-attention mechanism for efficient image encoding. 
In addition, we offer the option of using \textit{standard CNNs} on raw images for policy training from scratch.

{
\tiny
\begin{table}[t]
\centering
\begin{tabular}{@{} lcccc @{}}

\toprule

\textbf{Algorithms\textbackslash Tasks} & \textbf{Spread-4} & \textbf{Spread-5} & \textbf{Spread-8} \\

\midrule

\textbf{QMIX} & \(-1278.26 \pm 23.13\) & \(-2531.17 \pm 586.56\) & \(-6414.48 \pm 27.27\) & \\
\textbf{QPLEX} & \(\textbf{-766.84} \pm \textbf{14.39}\) & \(-1800.53 \pm 194.49\) & \(-13260.36 \pm 6200.03\) \\
\textbf{MAA2C} & \(-1190.09 \pm 99.93\) & \(-2312.56 \pm 222.08\) & \(\textbf{-5961.67} \pm \textbf{66.04}\) \\
\cellcolor{green!25}\textbf{MAPPO} & \(-971.17 \pm 124.22\) & \(-1910.20 \pm 42.86\) & \(\textbf{-5926.39} \pm \textbf{38.48}\) \\
\textbf{HAPPO} & \(-1032.80 \pm 45.84\) & \(-2000.41 \pm 98.20\) & \(-6940.61 \pm 69.55\) \\
\textbf{MAT-DEC} & \(-1066.62 \pm 45.98\) & \(-1918.88 \pm 15.76\) & \(-6843.44 \pm 563.39\) \\
\textbf{COMA} & \(-1176.78 \pm 33.37\) & \(-2003.47 \pm 51.18\) & \(-6249.07 \pm 44.73\) \\
\textbf{EOI} & \(-1963.23 \pm 859.27\) & \(-5816.66 \pm 20.34\) & \(-13210.98 \pm 5659.53\) \\
\textbf{MASER} & \(-969.06 \pm 5.01\) & \(-1939.58 \pm 113.22\) & \(-6242.61 \pm 14.40\) \\
\textbf{EMC} & \(-1216.19 \pm 10.9\) & \(-1961.75 \pm 0.71\) & \(-6219.14 \pm 29.55\) \\
\cellcolor{green!25}\textbf{CDS} & \(-809.20 \pm 47.22\) & \(\textbf{-1641.77} \pm \textbf{14.61}\) & \(-6250.72 \pm 15.44\) \\

\bottomrule

\end{tabular}
\caption{Results in \textit{Spread} (MPE) tasks.}
\label{tab:MPE-results}

\end{table}
}

{\tiny
\begin{table}[t]
\centering
\begin{tabular}{@{} lccc @{}}

\toprule

\textbf{Algorithms\textbackslash Envs} & \textbf{Pistonball} & \textbf{Cooperative Pong} & \textbf{Entompted Cooperative} \\

\midrule

\cellcolor{green!25}\textbf{QMIX} & \(\textbf{991.59} \pm \textbf{0.17}\) & \(\textbf{199.57} \pm \textbf{0.61}\) & \(8.00 \pm 0.00\) \\
\cellcolor{green!25}\textbf{QPLEX} & \(\textbf{991.46} \pm \textbf{0.10}\) & \(\textbf{197.78} \pm \textbf{3.14}\) & \(8.00 \pm 0.00\) \\
\textbf{MAA2C} & \(\textbf{990.83} \pm \textbf{0.04}\) & \(-0.33 \pm 3.47\) & \(6.57 \pm 0.05\) \\
\textbf{MAPPO} & \(\textbf{990.71} \pm \textbf{0.10}\) & \(13.22 \pm 4.61\) & \(6.61 \pm 0.05\) \\
\textbf{HAPPO} & \(983.60 \pm 0.77\) & \(25.62 \pm 3.36\) & \(7.99 \pm 0.01\) \\
\textbf{MAT-DEC} & \(982.57 \pm 2.44\) & \(\textbf{200.00} \pm \textbf{0.00}\) & \(10.00 \pm 0.00\) \\
\textbf{COMA} & \(678.28 \pm 324.06\) & \(1.10 \pm 7.51\) & \(7.68 \pm 1.7\) \\
\textbf{EOI} & \(948.35 \pm 42.74\) & \(-1.64 \pm 2.66\) & \(6.53 \pm 0.02\) \\
\textbf{MASER} & \(\textbf{989.39} \pm \textbf{0.64}\) & \(104.25 \pm 69.29\) & \(8.00 \pm 0.00\) \\
\textbf{EMC} & \(265.00 \pm 174.58\) & \(\textbf{196.5} \pm \textbf{2.83}\) & \(8.00 \pm 0.00\) \\
\textbf{CDS} & \(415.82 \pm 284.37\) & \(\textbf{197.13} \pm \textbf{2.26}\) & \(\textbf{11.10} \pm \textbf{1.00}\) \\

\bottomrule

\end{tabular}
\caption{Results in PettingZoo using ResNet18.}
\label{tab:pettingzoo-results}

\end{table}
}

\section{Benchmark Analysis}

\subsection{Experimental Setup}
In our benchmark analysis, we evaluate MARL algorithms on \textit{Entombed Cooperative}, \textit{Pistonball}, \textit{Cooperative Pong}, \textit{Cramped Room}, \textit{Asymmetric Advantages}, \textit{Coordination Ring}, \textit{Pressure Plate}, \textit{Spread}, \textit{LBF} and \textit{RWARE} benchmarks. To ensure a fair comparison of the selected algorithms, we utilize the same number of training timesteps. Specifically, we use 10 million timesteps for MPE and LBF, 40 million for RWARE, 5 million for PettingZoo; 20 million for PressurePlate, 40 million for the Overcooked's \textit{Cramped Room}, and 100 million timesteps for the Overcooked's \textit{Asymmetric Advantages} and \textit{Coordination Ring}. We note that for all PettingZoo tasks, where the observations are RGB images, we have used a pre-trained ResNet18 model as the observation image encoder. Moreover, except HAPPO, we utilize agents to share their policy parameters.

We adopt the experimental setup of~\cite{papoudakis2021benchmarking}: Throughout the training of all algorithms, we conducted 100 test episodes at every 50,000-step interval to evaluate the current policy. During these test episodes, we record the episode returns, i.e., the accumulated rewards per episode, and compute the average return $\frac{1}{N} \sum_{i=1}^N R_{t,i,j}$, where \( N = 100 \) is the number of test episodes, and $R_{t,i,j}$ is the return of the $i$-th test episode at timestep $t$ of the $j$-th experiment. 

{\tiny
\begin{table}[t]
\centering
\begin{tabular}{@{} lccc @{}}

\toprule

\textbf{Algorithms\textbackslash Envs} & \textbf{Cramped Room} & \textbf{Asymmetric Advantages} & \textbf{Coordination Ring} \\

\midrule

\textbf{QMIX} & \(0.00 \pm 0.00\) & \(300.00 \pm 300.00\) & \(0.00 \pm 0.00\) \\
\textbf{QPLEX} & \(86.67 \pm 122.57\) & \(0.00 \pm 0.00\) & \(0.00 \pm 0.00\) \\
\cellcolor{green!25}\textbf{MAA2C} & \(\textbf{286.80} \pm \textbf{9.34}\) & \(\textbf{487.80} \pm \textbf{107.60}\) & \(\textbf{0.10} \pm \textbf{0.10}\) \\
\textbf{MAPPO} & \(\textbf{280.00} \pm \textbf{0.00}\) & \(0.30 \pm 0.10\) & \(\textbf{0.07} \pm \textbf{0.09}\) \\
\textbf{HAPPO} & \(0.00 \pm 0.00\) & \(160.10 \pm 159.9\) & \(0.00 \pm 0.00\) \\
\textbf{MAT-DEC} & \(0.00 \pm 0.00\) & \(0.00 \pm 0.00\) & \(0.00 \pm 0.00\) \\
\textbf{COMA} & \(0.20 \pm 0.16\) & \(0.10 \pm 0.10\) & \(\textbf{0.07} \pm \textbf{0.09}\) \\
\textbf{EOI} & \(\textbf{280.0} \pm \textbf{0.00}\) & \(1.60 \pm 0.60\) & \(\textbf{0.13} \pm \textbf{0.09}\) \\
\textbf{EMC} & \(0.00 \pm 0.00\) & \(-\) & \(-\) \\
\textbf{MASER} & \(0.00 \pm 0.00\) & \(0.00 \pm 0.00\) & \(0.00 \pm 0.00\) \\
\textbf{CDS} & \(186.67 \pm 133.00\) & \(70.00 \pm 70.00\) & \(0.00 \pm 0.00\) \\

\bottomrule

\end{tabular}
\caption{Results in Overcooked environments.}
\label{tab:overcooked-results}
\end{table}
}

{\tiny
\begin{table}[t]
\centering
\begin{tabular}{@{} lcc @{}}

\toprule

\textbf{Algorithms\textbackslash Envs} & \textbf{4p} & \textbf{6p} \\

\midrule

\textbf{QMIX} & \(-210.72 \pm 17.84\) & \(-3461.77 \pm 1020.33\) \\
\textbf{QPLEX} & \(-652.19 \pm 9.29\) & \(-5183.56 \pm 345.46\) \\
\textbf{MAA2C} & \(-281.59 \pm 201.77\) & \(-547.39 \pm 21.00\) \\
\cellcolor{green!25}\textbf{MAPPO} & \(-135.99 \pm 1.32\) & \(\textbf{-494.08} \pm \textbf{10.54}\) \\
\textbf{HAPPO} & \(-258.69 \pm 224.93\) & \(\textbf{-584.90} \pm \textbf{201.68}\) \\
\textbf{MAT-DEC} & \(-876.58 \pm 1113.53\) & \(-2930.77 \pm 3652.53\) \\
\textbf{COMA} & \(-4391.79 \pm 108.96\) & \(-12360.20 \pm 314.06\) \\
\textbf{EOI} & \(-3050.64 \pm 1125.83\) & \(-9221.16 \pm 4796.08\) \\
\textbf{MASER} & \(\textbf{-88.44} \pm \textbf{2.84}\) & \(-5257.08 \pm 4099.19\) \\
\textbf{EMC} & \(-4518.23 \pm 249.68\) & \(-12347.60 \pm 0.0\) \\
\textbf{CDS} & \(-1926.45 \pm 260.85\) & \(-8068.23 \pm 519.79\) \\

\bottomrule

\end{tabular}
\caption{Results in Pressure Plate tasks.}
\label{tab:pressureplate-results}

\end{table}
}

\begin{figure*}[t]
\centering

\setlength{\abovecaptionskip}{10pt} 

\begin{tabular}{c c c} 

    \includegraphics[width=0.19\textwidth]{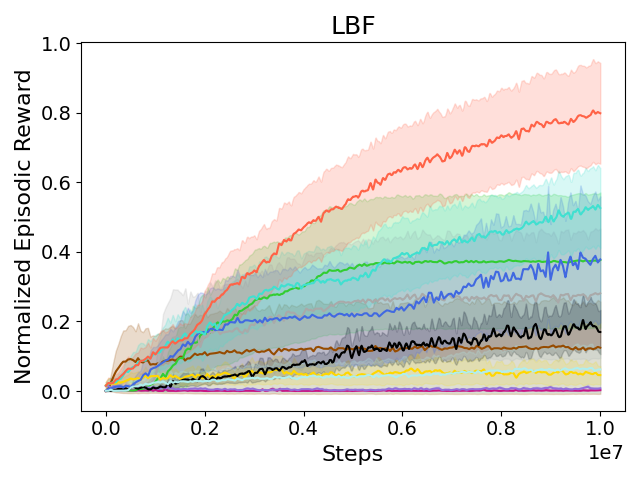} &
    \includegraphics[width=0.19\textwidth]{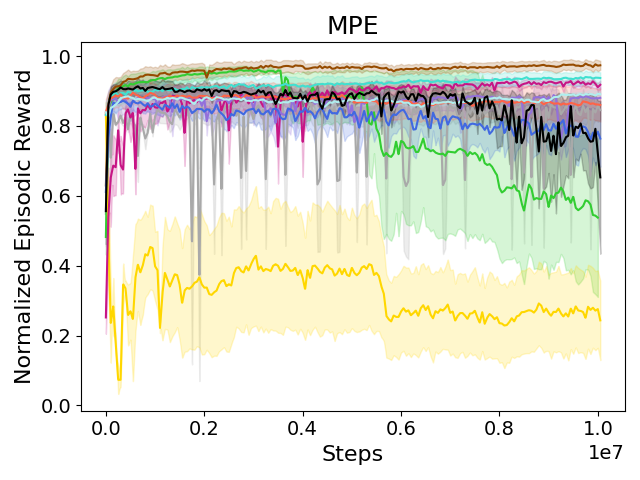} &
    \includegraphics[width=0.19\textwidth]{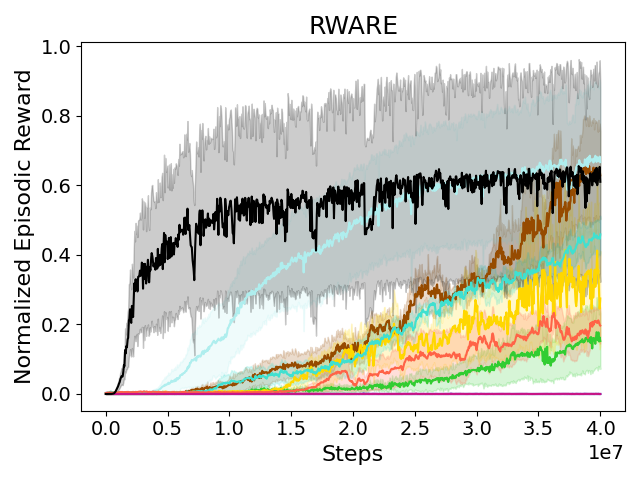} \\
    
    \includegraphics[width=0.19\textwidth]{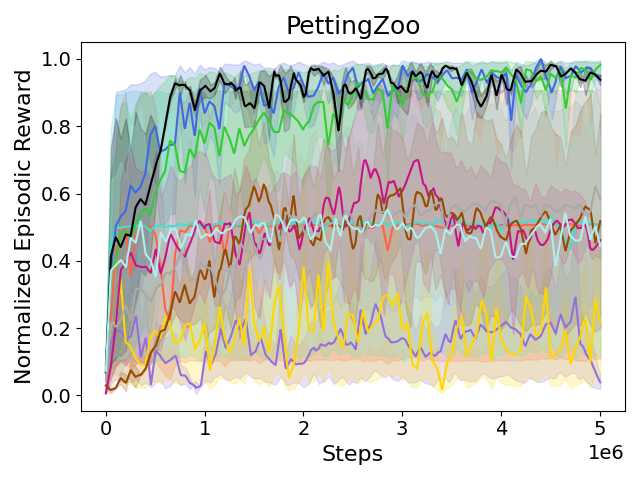} &
    \includegraphics[width=0.19\textwidth]{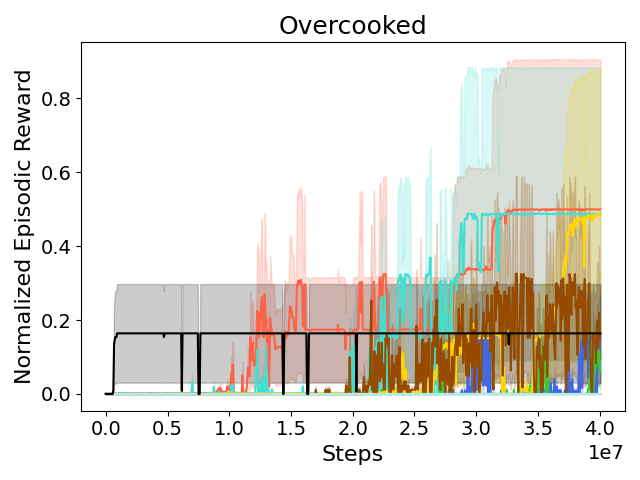} &
    \includegraphics[width=0.19\textwidth]{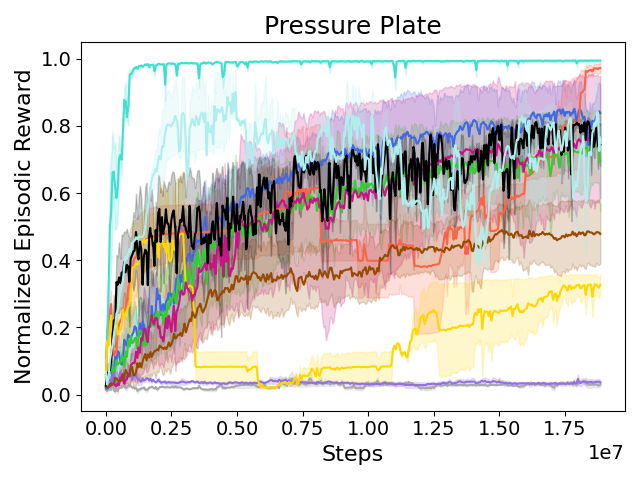} \\

    \multicolumn{3}{c}{\includegraphics[width=0.6\textwidth]{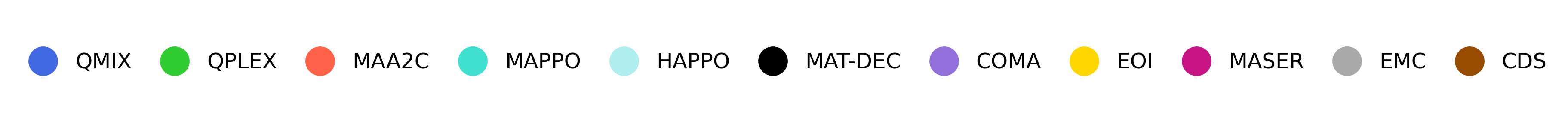}} \\

\end{tabular}

\caption{Aggregated Normalized episodic rewards of each benchmark.}
\label{fig:average_curves}
\end{figure*}

Our primary metric score is the mean of the unnormalized returns received from the best policy in convergence over five different seeds. As best policy in convergence, we use the policy that received the best return of the last 50 test episodes. 
We present our results using the mean and the 75$\%$ confidence interval.

We used the hyperparameter settings of the newly integrated algorithms, QPLEX, HAPPO, MAT-DEC, CDS, EOI, EMC, and MASER, adhering to configurations suggested by the authors of their original papers for challenging tasks. For MAA2C, MAPPO, COMA and QMIX, we employed the default parameters of EPyMARL. 
This ensures that the evaluation is as consistent as possible with what is reported in the original papers, allowing for a valid comparison between the algorithms and with results already reported. 
Our choice to forego extensive tuning is further supported by: {(a) preliminary results showing \textit{no significant differences} regarding algorithms' performance across tasks}, and (b) the fact that our results, are based on the mean returns from the \textit{best policy} across different seeds. 

All experiments were conducted using CPUs, except for tasks within the PettingZoo environment, where image inputs required the use of GPUs. For these image-based tasks, GPUs were utilized for both the frozen image-encoder and training CNN components when a pre-trained encoder was not available. Specifically, we employed two "g5.16xlarge" AWS instances, each featuring 64 vCPUs, 256 GB RAM, and one A10G GPU with 24 GB of memory to manage the computational demands of image processing.

\subsection{Main Results and Analysis}

In this section, we present our main results and the benchmark analysis. 
We show the main analytical results in Tables \ref{table:LBF-results}, \ref{tab:RWARE-results}, \ref{tab:MPE-results}, \ref{tab:pettingzoo-results}, \ref{tab:overcooked-results} and \ref{tab:pressureplate-results}, and the averaged main results in Figure \ref{fig:average_curves}. To obtain these average results, we use the \textit{normalized} scores for each task of the benchmark and average over these scores. In the tables, we write in bold the values of the best algorithm in a specific task. If the performance of another algorithm was not statistically significantly different from the best algorithm, the respective value is also in bold. Algorithms with green cell color are the best on average in the corresponding benchmark. 

{
\small
\begin{table}[t]
\centering
\begin{tabular}{@{} lccc @{}}

\toprule

\textbf{Algorithms} & \textbf{Episodic Reward} & \textbf{RAM} & \textbf{Training time} \\

\midrule

\textbf{MAA2C-ResNet18} & \(\textbf{990.83} \pm \textbf{0.04}\) & \(7GB\) & \(4d:15h\) \\
\textbf{MAA2C-CNN} & \(846.74 \pm 19.7\) & \textcolor{red}{\(34GB\)} & \textcolor{red}{\(16d:0h\)} \\

\bottomrule

\end{tabular}
\caption{Results in PettingZoo's Pistonball comparing ResNet18 as frozen image encoder with trainable CNNs.}
\label{tab:pettingzoo-results-ablation}

\end{table}
}

\begin{figure}[t]
    \centering
        \includegraphics[width=0.3\textwidth]{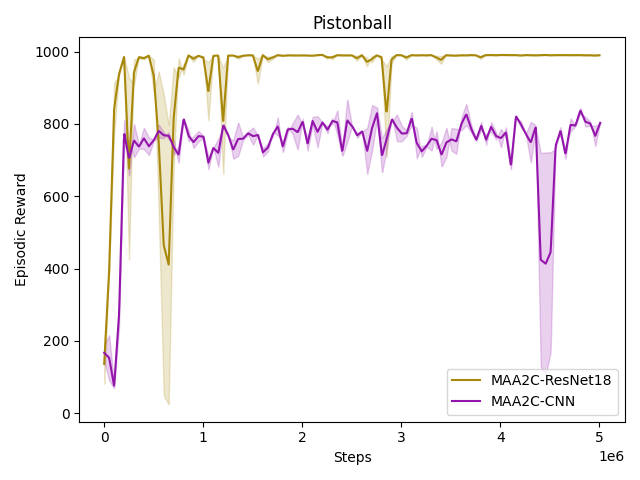}
    \caption{ResNet18 vs Trainable CNNs: MAA2C in PettingZoo's Pistonball task.}
    \label{fig:ablation}
\end{figure}

{
\tiny
\begin{table*}[t]
\centering

\begin{tabular}{@{} l *{11}{c} @{}}
\toprule
\textbf{Environments\textbackslash Algorithms} & \textbf{QMIX} & \textbf{QPLEX} & \textbf{MAA2C} & \textbf{MAPPO} & \textbf{HAPPO} & \textbf{MAT-DEC} & \textbf{COMA} & \textbf{EOI} & \textbf{MASER} & \textbf{EMC} & \textbf{CDS} \\
\midrule
\textbf{LBF} & \(0d:8h\) & \(0d:13h\) & \(0d:1h\) & \(0d:2h\) & \(0d:5h\) & \(0d:4h\) & \(0d:1h\) & \(0d:6h\) & \(0d:18h\) & \(2d:4h\) & \(0d:18h\) \\
\textbf{RWARE} & \(1d:22h\) & \(2d:17h\) & \(0d:9h\) & \(0d:12h\) & \(1d:1h\) & \(0d:20h\) & \(0d:9h\) & \(1d:18h\) & \(2d:16h\) & \(19d:1h\) & \(3d:2h\) \\
\textbf{Spread (MPE)} & \(0d:15h\) & \(0d:21h\) & \(0d:2h\) & \(0d:3h\) & \(0d:9h\) & \(0d:6h\) & \(0d:2h\) & \(0d:11h\) & \(1d:12h\) & \(4d:9h\) & \(4d:9h\) \\
\textbf{Petting Zoo} & \(1d:16h\) & \(3d:11h\) & \(3d:23h\) & \(3d:10h\) & \(14d:1h\) & \(3d:6h\) & \(3d:11h\) & \(0d:23h\) & \(2d:1h\) & \(3d:23h\) & \(1d:19h\) \\
\textbf{Overcooked} & \(3d:14h\) & \(5d:3h\) & \(0d:19h\) & \(1d:4h\) & \(1d:18h\) & \(2d:11h\) & \(0d:21h\) & \(3d:9h\) & \(4d:9h\) & \(13d:14h\) & \(2d:14h\) \\
\textbf{Pressure Plate} & \(0d:22h\) & \(1d:14h\) & \(0d:4h\) & \(0d:7h\) & \(0d:20h\) & \(0d:12h\) & \(0d:4h\) & \(0d:18h\) & \(1d:6h\) & \(9d:7h\) & \(2d:1h\) \\

\bottomrule
\end{tabular}
\caption{Average (wall-clock) training times over all tasks of each benchmark for all 11 algorithms.}
\label{tab:average_train_times_transposed}

\end{table*}
}

\subsubsection{Key Findings}

Based on results, we highlight the following:
\begin{itemize}
    \item {\color{black}\textbf{Best Algorithms.}} The standard MARL algorithms, QPLEX, MAPPO, MAA2C and CDS, demonstrate the most consistent performance across all fully cooperative benchmarks.
    \item {\color{black}\textbf{SoTA exploration-based algorithms mostly underperform.}} Exploration-based methods that have reached SoTA performance in SMAC and/or GRF, that is, EOI, EMC and MASER, significantly underperform in most benchmarks compared to the standard methods. In some cases, they even fail entirely, including in tasks with sparse reward settings (e.g., see Tables \ref{table:LBF-results}, \ref{tab:RWARE-results} and \ref{tab:overcooked-results}), in which these methods are expected to improve performance over baselines. The only exception is CDS which is shown to be one of the best algorithms in RWARE and Spread.  
    \item {\color{black}\textbf{Value Decomposition.}} As demonstrated by the results of QMIX and QPLEX, value decomposition methods tend to converge to suboptimal policies as the number of agents increases, significantly underperforming compared to actor-critic methods (e.g., see the results in Spread, LBF and Pressure Plate). Moreover, our findings contrast with one conclusion of \cite{papoudakis2021benchmarking} that suggests that value decomposition methods require sufficiently dense rewards to effectively learn to decompose the value function. We show several cases with sparse reward settings, such as in LBF (e.g., \textit{2s-8x8-3p-2f}, \textit{2s-9x9-3p-2f}), RWARE (\textit{tiny-4ag-hard}), and Pressure Plate (4p), where QPLEX, or even QMIX, can successfully converge to effective policies. 
\end{itemize}
 
\subsubsection{Discussion on Algorithms' Performance}

Next, we analyse the performance of the evaluated MARL algorithms in the selected fully cooperative benchmark tasks.

\paragraph{\textbf{QMIX}, \textbf{QPLEX}} 

QMIX shows mediocre performance across most tasks, with complete failure in some (e.g., RWARE and many LBF tasks), yet the highest rewards in  PettingZoo. Notably, QMIX is the only method to achieve positive rewards in the sparse-reward LBF \textit{4s-11x11-3p-2f} task. In contrast, QPLEX generally outperforms QMIX improving performance in cooperative tasks. However, QPLEX fails in the most challenging LBF and Overcooked tasks and struggles in large-scale scenarios such as \textit{Spread-8} and Pressure Plate's \textit{6p}. Interestingly, despite lacking advanced exploration techniques, QPLEX notably improves over QMIX in RWARE.

\paragraph{\textbf{MAA2C}, \textbf{MAPPO}, \textbf{COMA}} 
MAA2C is one of the most consistent algorithms across all tasks, showing the best performance in complex sparse-reward LBF tasks and in  \textit{Spread-8} (along with MAPPO) and Overcooked. Similarly, MAPPO also ranks as one of the most consistent algorithms but significantly outperforms MAA2C in RWARE, \textit{Spread-5}, \textit{Cooperative Pong}, and Pressure Plate's \textit{6p}. MAPPO achieves the highest rewards on average in Spread (together with CDS) and Pressure Plate. Interestingly, in the RWARE tasks where MAPPO had been the best option in previous work \cite{papoudakis2021benchmarking}, HAPPO, MAT-DEC, and CDS outperform it. Conversely, COMA has one of the lowest performances overall, with competitive results only in \textit{Spread}, \textit{Entombed Cooperative}, and \textit{Coordination Ring}. However, for the last two tasks, all algorithms perform poorly, so none are truly competitive.

\paragraph{\textbf{HAPPO}, \textbf{MAT-DEC}}

Both HAPPO and MAT-DEC do not manage to achieve consistent performance across the evaluated benchmarks. More specifically, both methods are shown to be among the best (together with CDS) in RWARE, while HAPPO also achieving good performance in Pressure Plate and MAT-DEC in the high-dimensional PettingZoo tasks. However, both HAPPO and MAT-DEC perform poorly in tasks where efficient exploration and synchronized coordination are crucial, such as in LBF and Overcooked benchmarks. 

\paragraph{\textbf{MASER}, \textbf{EMC}} 

Despite MASER and EMC achieving strong results in sparse-reward SMAC tasks, they perform poorly in most fully cooperative tasks. The notable exception is PettingZoo tasks, where MASER’s baseline, QMIX, performs better. MASER also achieves the highest rewards in Pressure Plate (\textit{4p}). Both MASER's and EMC's poor overall performance is attributed to its reliance on Q-values to enhance joint exploration: Since agents rarely receive positive rewards, this leads to misleading intrinsic rewards based on irrelevant Q-value-driven sub-goals. However, Q-values prove to be useful for guiding exploration in LBF's \textit{2s-12x12-2p-2f}, as this task is less sparse but also requires good exploration for success.

\paragraph{\textbf{EOI, CDS}}
EOI and CDS outperform MASER and EMC in most tasks. EOI achieves the highest rewards in Overcooked's \textit{Cramped Room} (alongside MAA2C and MAPPO) and RWARE's \textit{tiny-2ag-hard}. However, EOI lacks the consistency of the MAA2C backbone algorithm. We attribute this inconsistency to the emphasis on individuality, which can hinder full cooperation and the high level of coordination required in most fully cooperative tasks. In contrast, CDS is more consistent, showing top performance in Spread and RWARE (alongside MAT-DEC and HAPPO) tasks. CDS' improved performance is attributed to the fact that although CDS relies on encouraging agents to be more diverse, as EOI does, it also assures sufficient sharing of the most useful experience of the agents.

\subsubsection{Pre-trained Image-Encoder vs Trainable CNNs for Image-based Observations}
Our experimental results underscore the effectiveness of employing a frozen pre-trained image encoder, such as ResNet18, over trainable CNNs in multi-agent reinforcement learning (MARL) environments. In the specific case of the PettingZoo's Pistonball task, the use of frozen ResNet18 led to more stable and higher overall policy performance compared to its trainable counterparts. This advantage is clearly illustrated by the smoother convergence curves and the consistently superior performance throughout the training steps, as shown in Figure \ref{fig:ablation}. Moreover, as detailed in Table \ref{tab:pettingzoo-results-ablation}, the non-adaptive nature of frozen ResNet18 highlights its efficiency in managing complex visual contexts without computational overhead. This suggests that pre-trained, static encoders are highly beneficial, providing immediate and reliable performance improvements. The adoption of frozen image encoders significantly improves computational efficiency by eliminating the need for ongoing adjustments during training. This leads to shorter training times for MARL algorithms, as detailed in Table \ref{tab:pettingzoo-results-ablation}. Moreover, by converting image data into compact vector representations, these encoders substantially lower memory requirements, facilitating the execution of complex MARL tasks with RGB-based observations. 

\subsubsection{Comparison of training times}

As can be clearly seen from Table \ref{tab:average_train_times_transposed}, off-policy algorithms generally require longer training due to their use of large replay buffers, which process experiences from multiple past episodes. In contrast, on-policy algorithm, benefit from learning directly from current policy experiences, allowing for faster training. However, in image-based, high-dimensional environments, such as PettingZoo tasks, this parallelism can slow down training, as it does not integrate well with pre-trained models, such as ResNet18, which require GPU resources.

\subsubsection{Open Challenges} 

Based on the results we discussed above, it is evident that fully cooperative MARL tasks need careful algorithmic design, in terms of both excessive coordination and joint exploration, as current SoTA and standard MARL methods are not very effective. Below, we report some significant open challenges arised from our benchmark analysis:
\\ \textbf{(a)} The tasks \textit{Entombed Cooperative} (PettingZoo) and \textit{Coordination Ring} (Overcooked) are the most challenging, as all evaluated algorithms indeed fail to find any effective policy. Any improvement on these tasks would be of remarkable interest. 
\\ \textbf{(b)} The sparse-reward LBF tasks, with a large grid, three agents and two foods, the sparse-reward Pressure Plate tasks, with more than 4 agents, and the sparse-reward hard RWARE tasks, with larger grids, are quite challenging, as they require excessive coordinated exploration, and any improvement on these is very interesting.
\\ \textbf{(c)} The Spread tasks, with more than 4 agents are quite challenging, as they require excessive coordination, and any improvement on these without the use of agent communication during execution is very interesting.

\section{Related Work}
The recent rise in MARL popularity has fragmented community standards and tools, with the frequent introduction of new libraries such as \cite{rutherford2024jaxmarl, hu2022marllib, liang2018rllib, pettingzoo}. Among the most popular are PyMARL \cite{smac} and EPyMARL \cite{papoudakis2021benchmarking}, both of which have played a crucial role in driving the influx of cooperative MARL algorithms. However, these libraries have somewhat overlooked the integration of fully cooperative MARL environments, pushing researchers to focus on specific benchmarks, such as SMAC \cite{smac, smac2} and GRF \cite{grf}, raising concerns about the reliability and generalizability of the proposed algorithms \cite{protocol}. Despite recent efforts \cite{papoudakis2021benchmarking, yu2020benchmarking, bettini2024benchmarl, hu2022marllib} aiming to provide a comprehensive understanding of standard cooperative MARL algorithms through benchmarking, the evaluation of fully cooperative MARL still lacks systematic diversity and reliability.

\section{Conclusion}
In this paper, we highlight and address key concerns in the evaluation of cooperative MARL algorithms by providing an extended benchmarking of well-known MARL methods in fully cooperative tasks. Our extensive evaluations reveal significant discrepancies in the performance of SoTA methods, which can eventually underperform compared to standard baselines. Based on our analysis, as well as by open-sourcing PyMARLzoo+, this paper aims to motivate towards more systematic evaluation of MARL algorithms, encouraging broader adoption of diverse, fully cooperative benchmarks.



\begin{acks}
The research work contributed by George Papadopoulos was partially supported by the Hellenic Foundation for Research and Innovation (HFRI) under the 5th Call for HFRI PhD Fellowships (Fellowship Number: 20769). The authors also gratefully acknowledge GRNET – National Infrastructures for Research and Technology for providing AWS resources, and the University of Piraeus for funding the participation in the AAMAS 2025 conference and associated travel costs.
\end{acks}



\bibliographystyle{ACM-Reference-Format} 
\bibliography{sample}


\newpage

\appendix

\onecolumn

\section{Environment API}
\label{appendix:a}
The following Python script (\autoref{fig:env_api_example}) provides an example of executing an episode in Butterfly's {Pistonball} task of PettingZoo using random actions under our framework. To use a different environment, the user only needs to modify the \textit{args} variable accordingly. Below, we provide more detailed instructions. The script demonstrates: (a) importing necessary packages, (b) initializing the environment, and (c) running an episode by repeatedly rendering the environment and applying random actions until the episode ends.

\begin{figure}[H]
\centering
\begin{minipage}{0.48\textwidth}
\begin{lstlisting}[language=Python,]
# Import packages
from envs import REGISTRY as env_REGISTRY
import random as rnd
# Arguments for PettingZoo
args = {
  "env": "pettingzoo",
  "env_args": {
      "key": "pistonball_v6",
  }
}
# Initialize environment
env = env_REGISTRY[args["env"]](**args["env_args"])
n_agns = env.n_agents
n_acts = env.get_total_actions()
# Reset the environment
obs, state = env.reset()
done = False
# Run an episode
while not done:
    # Render the environment
    env.render()
    # Insert the policy's actions here
    actions = rnd.choices(range(0, n_acts), k=n_agns)
    # Apply an environment step
    reward, done, info = env.step(actions)
    obs = env.get_obs()
    state = env.get_state()
# Terminate the environment
env.close()
\end{lstlisting}
\end{minipage}
\caption{Indicative Python script for executing an episode in PettingZoo's Pistonball.}
\label{fig:env_api_example}
\end{figure}

\noindent Following the example presented above (\autoref{fig:env_api_example}), we provide the minimum \textit{args} required for each environment, as well as all of those that can be specified.
\\
\\
\noindent \textbf{PettingZoo \textit{args}}:

\begin{figure}[H]
\centering
\begin{minipage}{0.4\textwidth}
\begin{lstlisting}[language=Python, numbers=none]
args = {
  "env": "pettingzoo",
  "env_args": {
      "key": "pistonball_v6",
  }
}
\end{lstlisting}
\end{minipage}
\caption{Example of minimal \textit{args} for PettingZoo tasks.}
\label{fig:minimal_env_api_pettingzoo}
\end{figure}

\begin{figure}[H]
\centering
\begin{minipage}{0.4\textwidth}
\begin{lstlisting}[language=Python, numbers=none]
args = {
  "env": "pettingzoo",
  "env_args": {
      "key": "pistonball_v6",
      "seed": 1,
      "time_limit": 900,
      "partial_observation": False,
      "trainable_cnn": False,
      "image_encoder": "ResNet18",
      "image_encoder_batch_size": 10,
      "image_encoder_use_cuda": True,
      "centralized_image_encoding": True,
      "kwargs": ""
  }
}
\end{lstlisting}
\end{minipage}
\caption{Example of all \textit{args} for PettingZoo tasks.}
\label{fig:full_env_api_pettingzoo}
\end{figure}

where:
\begin{itemize}
\item \textit{key} is the selected PettingZoo task. Options: "pistonball\_v6", "cooperative\_pong\_v5", "entombed\_cooperative\_v3", "space\_invaders\_v2", "basketball\_pong\_v3", "boxing\_v2", "combat\_jet\_v1", "combat\_tank\_v3", "double\_dunk\_v3", "entombed\_competitive\_v3", "flag\_capture\_v2", "foozpong\_v3", "ice\_hockey\_v2", "joust\_v3", "mario\_bros\_v3", "maze\_craze\_v3", "othello\_v3", "pong\_v3", "quadrapong\_v4", "space\_war\_v2", "surround\_v2", "tennis\_v3", "video\_checkers\_v4", "volleyball\_pong\_v2", "warlords\_v3", "wizard\_of\_wor\_v3", "knights\_archers\_zombies\_v10", "chess\_v6", "connect\_four\_v3", "gin\_rummy\_v4", "go\_v5", "hanabi\_v5", "leduc\_holdem\_v4", "rps\_v2", "texas\_holdem\_no\_limit\_v6", "texas\_holdem\_v4", "tictactoe\_v3", "simple\_v3", "simple\_adversary\_v3", "simple\_crypto\_v3", "simple\_push\_v3", "simple\_reference\_v3", "simple\_speaker\_listener\_v4", "simple\_spread\_v3", "simple\_tag\_v3", "simple\_world\_comm\_v3", "multiwalker\_v9", "pursuit\_v4", "waterworld\_v4".

\item \textit{seed} specifies the random sequence of the first environment state which is helpful for reproducible evaluations. The default value is 1.

\item \textit{time\_limit} is the maximum number of steps before episode termination. The default value is 900 for all Butterfly tasks except from Pistonball, which is 125, and 10000 for all Atari tasks. For all SISL tasks the default value is 500, for all MPE tasks is 25, 

\item \textit{partial\_observation} defines whether to use partial observations or not. Only applicable in "space\_invaders\_v2" and "entombed\_cooperative\_v3". The default value is False.

\item \textit{trainable\_cnn} defines whether to encode the image observations using a pre-trained model or to return the raw images. The default value is False.

\item \textit{image\_encoder} is the selected image-encoder to use for encoding the image observations. Only applicable if \textit{trainable\_cnn} is True. Options: "ResNet18", "SlimSAM", "CLIP". The default value is "ResNet18".

\item \textit{image\_encoder\_batch\_size} is the number of images to encode at once with the selected image-encoder. Only applicable if \textit{trainable\_cnn} is True. The default value is 10.

\item \textit{image\_encoder\_use\_cuda} defines whether to use GPU or CPU for the selected image-encoder. In this way, the policies networks can be in a different device than the image-encoder. The default value is True, but it automatically turns to True if no GPU available is found.

\item \textit{centralized\_image\_encoding} defines whether to encode images in a centralized manner, thus avoiding to load as many image-encoders as the parallel environment processes, reducing the GPU memory required. However, this is only supported by our training API, where multiple parallel processes can be managed. When it is used only with the environment API, the effect is that the raw image observations are returned instead of the encoded representations.

\item \textit{kwargs} can consist of specific arguments supported by the selected task\footnote{All the arguments of each task can be found on the official page of PettingZoo: \url{https://pettingzoo.farama.org/}}. The user can provide the arguments in following format: "('arg1',arg1\_value), ('arg2',arg2\_value), ...". For instance, we can specify the number of pistons in the \textit{pistonball\_v6} task using: "('n\_pistons',4),". The default value is "", that is, no arguments provided.

\end{itemize}

\noindent \\ \textbf{Overcooked \textit{args}}:

\begin{figure}[H]
\centering
\begin{minipage}{0.4\textwidth}
\begin{lstlisting}[language=Python, numbers=none]
args = {
  "env": "overcooked",
  "env_args": {
      "key": "cramped_room",
  }
}
\end{lstlisting}
\end{minipage}
\caption{Example of minimal \textit{args} for Overcooked tasks.}
\label{fig:minimal_env_api_overcooked}
\end{figure}

\begin{figure}[H]
\centering
\begin{minipage}{0.4\textwidth}
\begin{lstlisting}[language=Python, numbers=none]
args = {
  "env": "overcooked",
  "env_args": {
      "key": "cramped_room",
      "seed": 1,
      "time_limit": 500,
      "reward_type": "sparse"
  }
}
\end{lstlisting}
\end{minipage}
\caption{Example of all \textit{args} for Overcooked tasks.}
\label{fig:full_env_api_overcooked}
\end{figure}

where:
\begin{itemize}

\item \textit{key} is the selected Overcooked scenario. Options: "cramped\_room", "asymmetric\_advantages", "coordination\_ring", "counter\_circuit", "forced\_coordination".

\item \textit{seed} has the same functionality as in PettingZoo. The default value is 1.

\item \textit{time\_limit} has the same functionality as in PettingZoo. The default value is 500.

\item \textit{reward\_type} defines whether to provide sparse rewards, that is, only when an order is ready, or shaped rewards, that is, each time a component of an order is ready. Options: "sparse", "shaped". The default value is "sparse".

\end{itemize}

\noindent \\ \textbf{Pressure Plate \textit{args}}:

\begin{figure}[H]
\centering
\begin{minipage}{0.4\textwidth}
\begin{lstlisting}[language=Python, numbers=none]
args = {
  "env": "pressureplate",
  "env_args": {
      "key": "pressureplate-linear-4p-v0",
  }
}
\end{lstlisting}
\end{minipage}
\caption{Example of minimal \textit{args} for Pressure Plate environment.}
\label{fig:minimal_env_api_pressureplate}
\end{figure}

\begin{figure}[H]
\centering
\begin{minipage}{0.4\textwidth}
\begin{lstlisting}[language=Python, numbers=none]
args = {
  "env": "pressureplate",
  "env_args": {
      "key": "pressureplate-linear-4p-v0",
      "seed": 1,
      "time_limit": 500,
  }
}
\end{lstlisting}
\end{minipage}
\caption{Example of all \textit{args} for Pressure Plate environment.}
\label{fig:full_env_api_pressureplate}
\end{figure}

where:
\begin{itemize}

\item \textit{key} is the selected Pressure Plate scenario. Options: "pressureplate-linear-4p-v0", "pressureplate-linear-5p-v0", "pressureplate-linear-6p-v0".

\item \textit{seed} has the same functionality as in PettingZoo. The default value is 1.

\item \textit{time\_limit} has the same functionality as in PettingZoo. The default value is 500.

\end{itemize}

\noindent \\ \textbf{Capture Target \textit{args}}:

\begin{figure}[H]
\centering
\begin{minipage}{0.4\textwidth}
\begin{lstlisting}[language=Python, numbers=none]
args = {
  "env": "capturetarget",
  "env_args": {
      "key": "CaptureTarget-6x6-1t-2a-v0",
  }
}
\end{lstlisting}
\end{minipage}
\caption{Example of minimal \textit{args} for Capture Target environment.}
\label{fig:minimal_env_api_capturetarget}
\end{figure}

\begin{figure}[H]
\centering
\begin{minipage}{0.4\textwidth}
\begin{lstlisting}[language=Python, numbers=none]
args = {
  "env": "capturetarget",
  "env_args": {
      "key": "CaptureTarget-6x6-1t-2a-v0",
      "seed": 1,
      "time_limit": 60,
      "obs_one_hot": False,
      "target_flick_prob": 0.3,
      "tgt_avoid_agent": True,
      "tgt_trans_noise": 0.0,
      "agent_trans_noise": 0.1,
  }
}
\end{lstlisting}
\end{minipage}
\caption{Example of all \textit{args} for Capture Target environment.}
\label{fig:full_env_api_capturetarget}
\end{figure}

where:
\begin{itemize}

\item \textit{key} is the selected Capture Target scenario. Options: "CaptureTarget-6x6-1t-2a-v0".

\item \textit{seed} has the same functionality as in PettingZoo. The default value is 1.

\item \textit{time\_limit} has the same functionality as in PettingZoo. The default value is 60.

\item \textit{obs\_one\_hot} defines whether to return the observations in one-hot encoding format. The default value is False.

\item \textit{target\_flick\_prob} specifies the probability of not observing the target. The default value is 0.3.

\item \textit{tgt\_avoid\_agent} defines whether the target keeps moving away from the agents or not. The default value is True. 

\item \textit{tgt\_trans\_noise} specifies the target's transition probability for arriving an unintended adjacent cell. The default value is 0.0.

\item \textit{agent\_trans\_noise} specifies the agent's transition probability for arriving an unintended adjacent cell. The default value is 0.1.

\end{itemize}
\noindent \\ \textbf{Box Pushing \textit{args}}:

\begin{figure}[H]
\centering
\begin{minipage}{0.4\textwidth}
\begin{lstlisting}[language=Python, numbers=none]
args = {
  "env": "boxpushing",
  "env_args": {
      "key": "BoxPushing-6x6-2a-v0",
  }
}
\end{lstlisting}
\end{minipage}
\caption{Example of minimal \textit{args} for Box pushing environment.}
\label{fig:minimal_env_api_boxpushing}
\end{figure}

\begin{figure}[H]
\centering
\begin{minipage}{0.4\textwidth}
\begin{lstlisting}[language=Python, numbers=none]
args = {
  "env": "boxpushing",
  "env_args": {
      "key": "BoxPushing-6x6-2a-v0",
      "seed": 1,
      "time_limit": 60,
      "random_init": True 
  }
}
\end{lstlisting}
\end{minipage}
\caption{Example of all \textit{args} for Box Pushing environment.}
\label{fig:full_env_api_boxpushing}
\end{figure}

where:
\begin{itemize}

\item \textit{key} is the selected Box Pushing scenario. Options: "BoxPushing-6x6-2a-v0".

\item \textit{seed} has the same functionality as in PettingZoo. The default value is 1.

\item \textit{time\_limit} has the same functionality as in PettingZoo. The default value is 60.

\item \textit{random\_init} shows whether random initial position of agents should be applied. The default value is True.
\end{itemize}

\noindent \\ \textbf{LBF, RWARE, MPE \textit{args}}:

\begin{figure}[H]
\centering
\begin{minipage}{0.4\textwidth}
\begin{lstlisting}[language=Python, numbers=none]
args = {
  "env": "gymma",
  "env_args": {
      "key": "mpe:SimpleSpeakerListener-v0",
  }
}
\end{lstlisting}
\end{minipage}
\caption{Example of minimal \textit{args} for LBF, RWARE, and MPE environments.}
\label{fig:minimal_env_api_gym}
\end{figure}

\begin{figure}[H]
\centering
\begin{minipage}{0.4\textwidth}
\begin{lstlisting}[language=Python, numbers=none]
args = {
  "env": "gymma",
  "env_args": {
      "key": "mpe:SimpleSpeakerListener-v0",
      "seed": 1,
      "time_limit": 500,
  }
}
\end{lstlisting}
\end{minipage}
\caption{Example of all \textit{args} for LBF, RWARE, and MPE environments.}
\label{fig:full_env_api_gymma}
\end{figure}

where:
\begin{itemize}

\item \textit{key} is the selected scenario of LBF, RWARE, or MPE environments. 
\\ Options for LBF: 
\\ "lbforaging:Foraging-4s-11x11-3p-2f-coop-v2", "lbforaging:Foraging-2s-11x11-3p-2f-coop-v2", "lbforaging:Foraging-2s-8x8-3p-2f-coop-v2", "lbforaging:Foraging-2s-9x9-3p-2f-coop-v2", "lbforaging:Foraging-7s-20x20-5p-3f-coop-v2", "lbforaging:Foraging-2s-12x12-2p-2f-coop-v2", "lbforaging:Foraging-8s-25x25-8p-5f-coop-v2", "lbforaging:Foraging-7s-30x30-7p-4f-coop-v2". 
\\ Options for RWARE: 
\\ "rware:rware-small-4ag-hard-v1", "rware:rware-tiny-4ag-hard-v1", "rware:rware-tiny-2ag-hard-v1".
\\ Options for MPE: 
\\ "mpe:SimpleSpread-3-v0", "mpe:SimpleSpread-4-v0", "mpe:SimpleSpread-5-v0", "mpe:SimpleSpread-8-v0", "mpe:SimpleSpeakerListener-v0", "mpe:MultiSpeakerListener-v0".

\item \textit{seed} has the same functionality as in PettingZoo. The default value is 1.

\item \textit{time\_limit} has the same functionality as in PettingZoo. The default value is 50 for LBF, 500 for RWARE, and 25 for MPE.

\end{itemize}
\section{Installation}
\label{appendix:b}
In this section, we provide the bash command needed for installing the PyMARLzoo+ environment below in \autoref{fig:min_commands}, assuming that Python >= 3.8 is installed on the machine. This minimal installation step is intended to facilitate a smooth setup process, thereby enabling users to access and utilize the framework's features efficiently and with ease.
\begin{figure}[htbp]
\centering
\centering
\begin{minipage}{0.4\textwidth}
\begin{lstlisting}[language=bash, numbers=none]
$ pip install pymarlzooplus
\end{lstlisting}
\end{minipage}
\caption{Bash command to install PyMARLzoo+ package using the pip tool.}
\label{fig:min_commands}
\end{figure}
\section{Training API}

The training framework of (E)PyMARL has been enhanced to support the integration of new algorithms and environments. We introduced a novel category of algorithm modules, termed \textit{explorers}, which integrate the exploration components of the CDS and EOI algorithms. This integration facilitates a structured and transparent code repository that accommodates both general-purpose and exploration-specific algorithms. Furthermore, the widely used \textit{Prioritized Replay Buffer} \cite{schaul2015prioritized} has been universally integrated into all off-policy algorithms, promoting efficient experimentation without necessitating code modifications. 
\autoref{fig:train_api_example} illustrates command examples for training an algorithm (\textit{<algo>}) on each environment task, demonstrating the minimal user effort required. The available options for each \textit{<algo>} and \textit{ <key>} and \textit{<time\_limit>} are described in \autoref{appendix:a}. Examples of usage of the extra arguments for PettingZoo and Overcooked are provided in \autoref{fig:extra_train_api_example}, in which 10 pistons and sparse reward type has been used for each environment, respectively.

\begin{figure}[htbp]
\raggedright
\textbf{LBF, RWARE, MPE}
\begin{lstlisting}[language=bash, numbers=none, basicstyle=\tiny]
$ python3 src/main.py --config=<algo> --env-config=gymma with env_args.time_limit=<time_limit> env_args.key=<key> 
\end{lstlisting}
\textbf{PettingZoo}
\begin{lstlisting}[language=bash, numbers=none, basicstyle=\tiny]
$ python3 src/main.py --config=<algo> --env-config=pettingzoo with env_args.time_limit=<time_limit> env_args.key=<key> env_args.kwargs=<kwargs>
\end{lstlisting}
\textbf{Overcooked}
\begin{lstlisting}[language=bash, numbers=none, basicstyle=\tiny]
$ python3 src/main.py --config=<algo> --env-config=overcooked with env_args.time_limit=<time_limit> env_args.key=<key> env_args.reward_type=<reward_type>
\end{lstlisting}
\textbf{PressurePlate}
\begin{lstlisting}[language=bash, numbers=none, basicstyle=\tiny]
$ python3 src/main.py --config=<algo> --env-config=pressureplate with env_args.key=<key> env_args.time_limit=<time_limit>
\end{lstlisting}
\textbf{Capture Target}
\begin{lstlisting}[language=bash, numbers=none, basicstyle=\tiny]
$ python3 src/main.py --config=<algo> --env-config=capturetarget with env_args.key=<key> env_args.time_limit=<time_limit>
\end{lstlisting}
\textbf{Box Pushing}
\begin{lstlisting}[language=bash, numbers=none, basicstyle=\tiny]
$ python3 src/main.py --config=<algo> --env-config=boxpushing with env_args.key=<key> env_args.time_limit=<time_limit>
\end{lstlisting}
\caption{Example Bash commands for training with different algorithms in different environment tasks. The \textit{<algo>}, and \textit{ <key>} correspond to the available options as mentioned in \autoref{appendix:a}. The \textit{<time\_limit>} field corresponds to the maximum number of steps before episode termination, therefore it should be an integer. }
\label{fig:train_api_example}
\end{figure}
\begin{figure}[htbp]
\raggedright
\textbf{PettingZoo} with 10 pistons in the Pistonball task for training QMIX with time limit 900 steps.
\begin{lstlisting}[language=bash, numbers=none, basicstyle=\tiny]
$ python3 src/main.py --config=qmix --env-config=pettingzoo with env_args.time_limit=900 env_args.key="pistonball_v6" env_args.kwargs="('n_pistons',10),"
\end{lstlisting}
\textbf{Overcooked} with sparse reward type in Cramped room scenario for training QMIX with time limit 900 steps.
\begin{lstlisting}[language=bash, numbers=none, basicstyle=\tiny]
$ python3 src/main.py --config=qmix --env-config=overcooked with env_args.time_limit=900 env_args.key="cramped_room" env_args.reward_type="sparse"
\end{lstlisting}

\caption{Example Bash commands for training QMIX algorithm in PettingZoo and Overcooked environment tasks.}
\label{fig:extra_train_api_example}
\end{figure}
\newpage

\section{Multi-Agent Reinforcement Learning Environments}\label{appendix:envs}
\label{appendix:c}
\subsection{Multi-Agent Particle Environment Details}
\begin{figure}[h]
    \centering
        \frame{\includegraphics[width=0.22\textwidth]{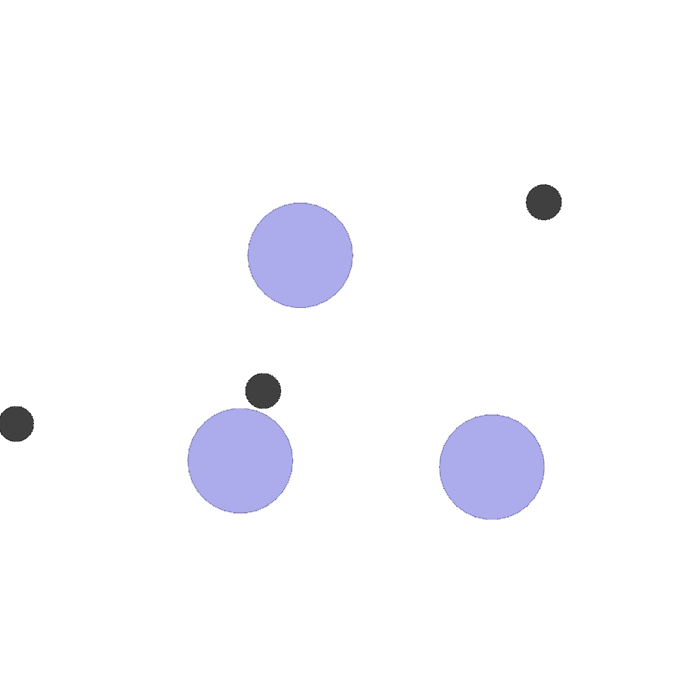} }
    \caption{Sample snapshot of the Spread MPE environment.}
    \label{fig:mpe}
\end{figure}

For our experiments in the MPE environment, we focused on the \textbf{Spread} environment, which corresponds to the cooperative navigation environment, as described in the paper by \citeauthor{mpe2} \cite{mpe2}. An example of the environment task is visualized in \autoref{fig:mpe}.

\subsubsection{Observations}\hfill\\



The environment involves N agents and N landmarks, where the agents are tasked with navigating to and covering the landmarks while preventing collisions with each other. Each agent observes its velocity and position, the relative position of other landmarks and agents, and lastly, the communication from all other agents. 
\[\begin{array}{c}

\text{O}^i = (velocity\_{x}, velocity\_{y}, agent\_position\_{x}, agent\_position\_y, 
N \text{×} (relative\_position\_x\_landmarks, relative\_position\_y\_landmarks), \\
(N-1) \text{×} (relative\_position\_x\_other\_agent, relative\_position\_y\_other\_agent) , (N+1) \text{×} one\_hot\_value )

\end{array}
\]

An example of an agent's observation space is provided below for the "SimpleSpread-3-v0" task:

\begin{lstlisting}[numbers=none]
array([-0.5         0.          0.8486883  -0.03767432 -0.44009647  0.24901201, 
       -1.0116129  -0.4476194  -1.7624141   0.89627826  0.09657926 -0.60416883, 
       -1.5796437  -0.8132916   0.          0.          0.          0.        ], dtype=float32),
\end{lstlisting}

This example observation vector consists of a vector, with 18 values, as it corresponds to N = 3 agents and landmarks. For N = 4 agents and landmarks, the observation vector would contain 24 values per agent; for N = 5, it would contain 30 values, and so on. 

\subsubsection{Actions}\hfill\\
The agent's action space includes the five standard actions that an agent $i$ can choose from:
$$A^{i} = (\text{No action, Move left, Move right, Move down, Move up})$$

\subsubsection{Rewards}\hfill\\
Each agent is awarded jointly based on the distance the agent is closest to every landmark. The distance is calculated based on the sum of minimum distances. When agents run into one another, they are penalized locally with -1 for each collision.

\subsection{Multi-Robot Warehouse (RWARE) Environment Details}
\begin{figure}[h!]
    \centering
        \frame{\includegraphics[width=0.15\textwidth]{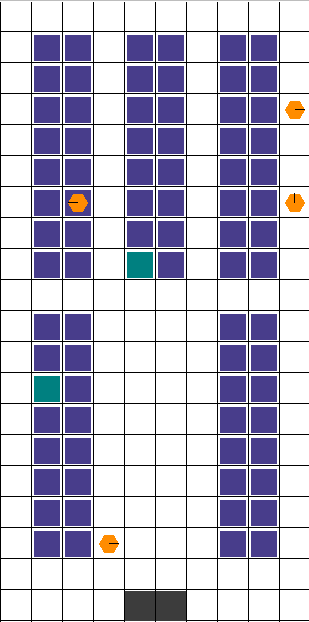} }
    \caption{Sample snapshot of small size RWARE environment with four agents.}
    \label{fig:rware}
\end{figure}

The Multi-Robot Warehouse (RWARE) environment \cite{papoudakis2021benchmarking}, designed based on real-world scenarios, depicts a warehouse that includes robots operating and gathering shelves, transferring them and the ordered items to a specific location, the workstation. Robots move the items back to empty shelf locations after humans have obtained the contents of a shelf. This cooperative environment has different settings that can be configured, like the size, the number of agents, the communication, and reward options. In our experiments, the tiny (10x11), and small (10x20) warehouse sizes were used, leveraging two and four agents on the 'hard' difficulty level. A sample snapshot of the "rware-small-4ag-hard-v1" task is given in \autoref{fig:rware}.
\subsubsection{Observations}\hfill\\
In this environment, each agent observation space is partially observable and includes information about the 3×3 grid, centered around itself, that an agent can observe. The information that is observed incorporates the agent’s own state (position, rotation, load) and the relative state of nearby agents and shelves. Specifically, the observation space for each agent $i$ can be described as a vector containing the following elements:

$$O^{i} = (\text{position\_x}, \text{position\_y}, \text{carrying\_shelf}, \text{direction\_one\_hot}, \text{path\_restricted}, \text{grid\_info})$$

where: 
\begin{itemize}
    \item \text{position\_x}, \text{position\_y}: The agent’s current position in x and y coordinates.
    \item \text{carrying\_shelf}: A binary value (1 or 0) showing whether the agent is carrying a shelf.
    \item \text{direction\_one\_hot}: Four values of the agent’s current direction (up, down, left, right) for each item in one-hot encoding.
    \item \text{path\_restricted}: A binary value (1 or 0) indicating if the agent is on a path where shelves cannot be placed.
    \item \text{grid\_info}: Information about the 3×3 grid surrounding the agent, split into 9 groups (one for each cell). The visibility range can be adjusted as well. Each group contains 7 elements:
    \begin{itemize}
    \item A binary value indicating if an agent exists in the cell.
\item Four values in one-hot encoding, representing the direction of any agent in the cell.
\item A binary value showing the presence of a shelf.
\item A binary value describing if the shelf is requested for delivery.
\end{itemize}
\end{itemize}

An example of an agent's observation array on the "rware-small-4ag-hard-v1" task with a 3×3 grid is provided below:
\begin{lstlisting}[numbers=none]
array([5. 5. 1. 1. 0. 0. 0. 0. 0. 1. 0. 0. 0. 1. 0. 0. 1. 0. 0. 0. 1. 0. 0. 1., 
       0. 0. 0. 0. 0. 0. 1. 0. 0. 0. 1. 0. 1. 1. 0. 0. 0. 1. 0. 0. 1. 0. 0. 0., 
       0. 0. 0. 1. 0. 0. 0. 1. 0. 0. 1. 0. 0. 0. 1. 0. 0. 1. 0. 0. 0. 0. 0.], dtype=float32),

\end{lstlisting}
\subsubsection{Actions}\hfill\\
The action space of this environment for each agent is comprised of four discrete possible options. For each agent i, the action space is the following: 

$$A^{i}=(\text{Turn left, Turn right, Forward, Load/Unload shelf})$$

Each agent can only rotate and move forward with the first three actions. The last action, relating to loading and unloading a shelf, can be performed only when a robot is in a specific location, below a shelf. 
\subsubsection{Rewards}\hfill\\
In every timestep, a fixed number of shelves (R) is requested to be transferred to a target location, and new shelves are randomly selected and included in the current requests. The reward value of an agent is 1 when it correctly delivers the ordered shelf. A difficulty that arises for the agents is to both complete the deliveries and find an empty spot to return the previously delivered shelf. Since multiple actions are needed between the shelf delivery by the agents, the reward signal is sparse. 

\subsection{PettingZoo Environment Details}
\begin{figure}[h!]
    \centering
    \begin{subfigure}[b]{0.22\textwidth}
        \centering
        \frame{\includegraphics[width=\textwidth]{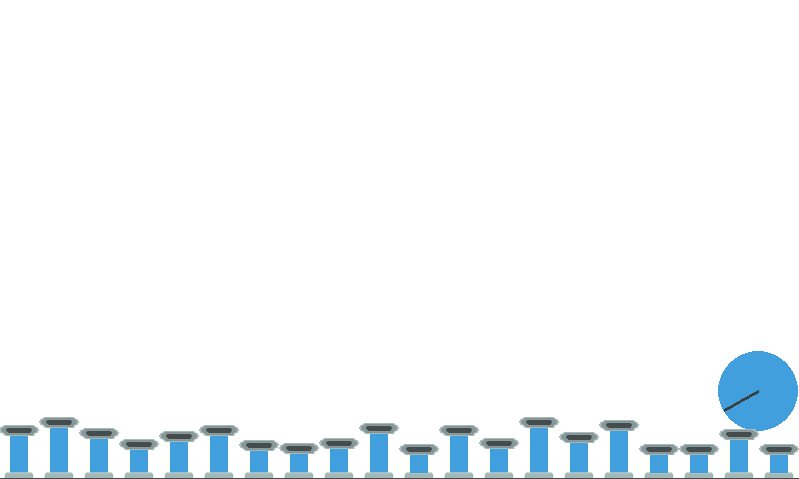}} 
        \caption{\label{fig:pistonball}Pistonball}
    \end{subfigure}
\quad
    \begin{subfigure}[b]{0.22\textwidth}
        \centering
        \frame{\includegraphics[width=\textwidth]{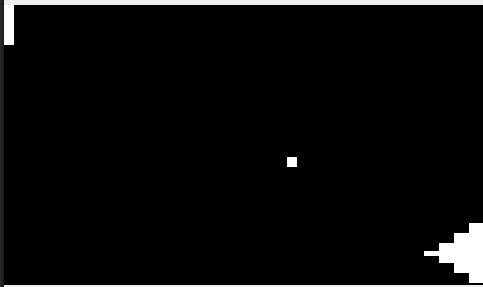} }
        \caption{\label{fig:pong}Cooperative Pong}
    \end{subfigure}
\quad
    \begin{subfigure}[b]{0.22\textwidth}
        \centering
        \frame{\includegraphics[width=0.7\textwidth]{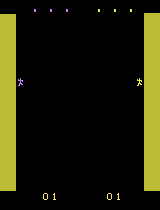} }
        \caption{\label{fig:entombed}Entombed Cooperative}
    \end{subfigure}
    \caption{Sample snapshots of the three PettingZoo game environments.}
    \label{fig:pettingzoo}
\end{figure}

For the PettingZoo environments\cite{pettingzoo}, we have mainly used the tasks from Atari and Butterfly environments, the  Entombed Cooperative, the Pistonball, and the Cooperative Pong. In the exploration game Entombed Cooperative, the agents must assist each other cooperatively to get through the evolving maze as far as they can. In the cooperative Pistonball physics game, the goal is to activate the pistons (agents) that move vertically so as to navigate the ball to the left wall of the game border. The Cooperative Pong game is a collaborative version of a pong game, where two agents (paddles) work together to continue playing with the ball for the longest period. Snapshot visualizations of these three fully cooperative environment games are illustrated in \autoref{fig:pettingzoo}.
\subsubsection{Observations}
\begin{figure}[h]    
        \centering
        \frame{\includegraphics[width=0.1\textwidth]{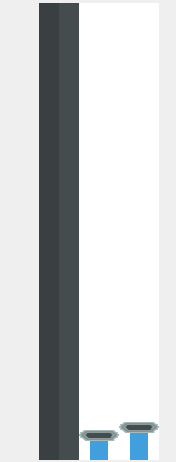} }
        
    \caption{Sample image-based observation of a piston in the Pistonball environment of PettingZoo with 10 agents.}
    \label{fig:pistonballobs}
\end{figure}

\hfill\\
\textbf{Pistonball}
The agent's observation space is composed of RBG values representing an image that illustrates the area occupied by the two pistons or the wall adjacent to the agent. 
An example of image-based observation of an agent (piston) can be found in \autoref{fig:pistonballobs} which is composed of RGB values and is of total size (457, 120, 3).
In case the image encoding technique is used, as described in the paper, the image observations are transformed into vectors. In that case, an example observation vector of a piston in the environment with 10 agents is given as a vector of size (512,) below: 

\begin{lstlisting}[numbers=none]
array([7.91885853e-01 0.00000000e+00 7.57015705e-01 7.02866971e-01, 
       1.37508601e-01 0.00000000e+00 9.95369107e-02 9.06493664e-02, 
       ... 
       1.01920377e-04 2.64243525e-03 2.42827028e-01 1.12921096e-01, 
       1.05303168e+00 3.09428126e-01 1.34141669e-01 2.01561041e-02],dtype=float32])
\end{lstlisting}

\textbf{Cooperative Pong}
The observation space of each agent in this game is the entire screen of the game. An example of image-based observation of an agent (paddle) can be seen in \autoref{fig:pong} which is composed of RGB values and is of total size (280, 480, 3).

In case the image encoding technique is utilised, the image observations are transformed into vectors. An example observation vector of one of the paddles in the environment is given as a vector of size (512,) below: 

\begin{lstlisting}[numbers=none]
array([4.37711984e-01 5.60766220e-01 1.59607649e+00 1.11085367e+00, 
       3.02959293e-01 3.26711655e-01 5.47930487e-02 9.35874805e-02, 
       ...
       3.57522756e-01 3.15122038e-01 3.19070041e-01 4.04478341e-01, 
       9.99535978e-01 1.10822819e-01 2.30887890e-01 4.81555223e-01],dtype=float32])
\end{lstlisting}

\begin{figure}[h!]
    \centering
    \begin{subfigure}[b]{0.22\textwidth}
        \centering
        \frame{\includegraphics[width=\textwidth]{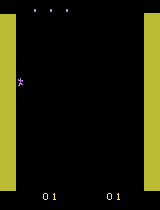}} 
        \caption{\label{fig:agent1}Agent 1}
    \end{subfigure}
\quad
    \begin{subfigure}[b]{0.22\textwidth}
        \centering
        \frame{\includegraphics[width=\textwidth]{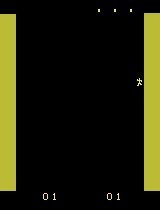} }
        \caption{\label{fig:agent2}Agent 2}
    \end{subfigure}
    \caption{Sample image-based partial observations of each agent in the Entombed Cooperative environment of PettingZoo.}
    \label{fig:entombedobs}
\end{figure}

\textbf{Entombed Cooperative}
In this game, each agent's observations are the RGB values - the entire screen of the game. An example of image-based partial observations of the two agents can be seen in \autoref{fig:entombedobs} which are composed of RGB values and are of total size (210, 160, 3).

In case the image encoding technique is utilized, the image observations are transformed into vectors. An example observation vector of one of the agents in the environment is given as a vector of size (512,) below: 

\begin{lstlisting}[numbers=none]
array([1.31586099e+00 4.78404522e-01 1.17896628e+00 4.77674603e-01, 
       2.60396451e-01 8.15561116e-02 2.70262837e-01 3.13673377e-01, 
       ...
       1.10996671e-01 1.78684831e-01 7.65883446e-01 8.93630207e-01, 
       1.32799280e+00 8.94335330e-01 6.74108326e-01 4.32223547e-03],dtype=float32])

\end{lstlisting}

Note that in our enhanced version, we implemented a method to mitigate flickering issues, which involves combining images from sequential frames. Specifically, our wrapper merges images from two consecutive frames to stabilize the representation of agent positions, thereby addressing the flickering problem. This process is repeated for four sequential steps to ensure consistency across frames. Our method identifies the position of Agent 1 in each frame by detecting specific RGB values and determines which frame depicts the agent more prominently. It then creates a combined image where the confirmed position of Agent 1 is reinforced. Additionally, for scenarios requiring partial observations, our wrapper generates separate observations for each agent by removing the visual representation of the other agent from the combined image. This approach not only stabilizes visual inputs but also supports scenarios where agents receive only their individual perspectives, enhancing the task's applicability for studies on cooperative perception and action under partial observability conditions.

\subsubsection{Actions}\hfill\\
\textbf{Pistonball}
The actions can be given either in discrete or continuous mode. In our experiments, the discrete action space has been used in which three actions are available for each agent i:
$$A^{i}=(\text{Move down, Stay still, Move up})$$

If the actions are continuous, then the range value is [-1, 1], scaled by four, and is linked to how much the pistons are moved up or down. In that way, action 1 will raise a piston by four pixels and -1 lower by four pixels.

\textbf{Cooperative Pong}
The discrete action space of each agent is composed by two actions. Each agent can be moved either up or down. Therefore, the action space for each agent $i$ is the following: 
$$A^{i}=(\text{Move up, Move down})$$

\textbf{Entombed Cooperative}
The action space of each agent is composed of 18 options. In an agent's turn, the actions that can be taken are the following:
\[A^{i}=(\text{No operation, Fire, Move up, Move right, Move left, Move down, Move upright, Move up left, Move downright, Move down left,}\]
\[\text{Fire up, Fire right, Fire left, Fire down, Fire up right, Fire up left, Fire down right, Fire down left})\]

\subsubsection{Rewards}\hfill\\
\textbf{Pistonball}
 The reward of each agent is composed of the overall amount of leftward movement of the ball and the leftward movement if it was close to the piston. In that way, each piston can take a global reward, while also a local reward is applied to those agents that are close to the ball, meaning positioned under any area of the ball. The local reward is computed as:

$$ \text{local\_reward} = 0.5 \times \Delta x_{\text{ball}}$$

where $\Delta x_{\text{ball}}$ represents the change in the ball’s x-position. The global reward is calculated as:

$$\text{global\_reward} = \left(\frac{\Delta x_{\text{ball}}}{x_{\text{start}}}\right) \times 100 + \text{time\_penalty}$$

where $x_{\text{start}}$ is the starting position of the ball and $\text{time\_penalty}$ is a default value of -0.1. Overall, the reward for each agent is calculated as :

$$R^{i} = \text{local\_{ratio}} * \text{local\_{reward}} + (1- \text{local\_{ratio}}) * \text{global\_reward}. $$

\textbf{Cooperative Pong}
At each time step, the reward for each agent $i$ when the ball remains successfully within the bounds is calculated as: \[ \text{R}^{i} = \frac{\text{max\_reward}}{\text{max\_cycles}} \] where the default value is set to 0.11. Alternatively, if the ball goes out of bounds, the agent receives a penalty determined by the parameter \text{off\_screen\_penalty}, which by default is set to -10. Upon this event, the game terminates.

\textbf{Entombed Cooperative}
The reward, identical to every single agent, in this environment, is given immediately after altering a section or resetting a stage, where each stage is composed of 5 sections. The reward could be received only when an agent loses a life, but not when it loses the last life as the game ends without the option of stage resetting.

\subsection{Level-Based Foraging (LBF) Environment Details}
The Level-Based Foraging \cite{lbf} environment is a multi-agent reinforcement learning environment built on OpenAI's RL framework \cite{1606.01540}. It features multiple agents operating within a grid-world setting, where their shared objective is to collect as much food as possible. The success of each agent in picking up food is governed by its level of compatibility with the task. The nature of the agents can be either competitive or cooperative in order to reach the goal. Success requires a balance of both to achieve optimal rewards.
For our experiments, we analyzed different configurations, varying the amount of players, food, grid size as well as observation radius.

\subsubsection{Observations}\hfill\\
In this environment, each agent has to navigate to the food scattered in the grid-world map and collect as many as possible. However, an agent can only pick up one food item, if its level is less than or equal to the combined levels of all collaborating agents. The game is succeeded, if all such food items are collected.
Each agent has knowledge either the full state or, if specified, a partial state of the environment. This includes the positions and levels of all entities in the map. The observation array contains N subarrays, each being the observation array of agent $i$ respectfully.
The observation array of each agent $i$ contains info on all existing food instances and agents, including themselves. The info for each observed instance is a triplet of the form: 
\[O^{i} = (\text{absolute\_position\_x}, \text{absolute\_position\_y}, \text{level})
\]
The final observation space for each agent is of size: $(\text{x}, \text{y}, \text{level}) \times \text{food\_count} + (\text{x}, \text{y}, \text{level}) \times \text{player\_count} = 3 \times (\text{food\_count} + \text{player\_count})$

If a food is picked up or another player is not found yet, the triplet value for that object is set as (-1, -1, 0 ).

Below is an example of an observation vector of an agent, for the environment "Foraging-4s-11x11-3p-2f-coop-v2":
\begin{lstlisting}[numbers=none]
array([ 2.,  0.,  4., -1., -1.,  0.,  2.,  4.,  2.,            
        2.,  3.,  1., -1., -1.,  0.], dtype=float32), 
\end{lstlisting}
 The three values in the array correspond to the triplet (position\_x, position\_y, level) of the first food item on the map, as observed by the agent. In this specific configuration, the environment consists of 2 food items and 3 players. Therefore, the first two triplets correspond to the food, while the rest represent the players. The second triplet of the array in particular is out of sight, as indicated by the initialized empty values (-1, -1, 0).
\subsubsection{Actions}\hfill\\
For this environment, the agent can perform 6 discrete actions. At each time step, the environment returns an action list, which contains all actions taken by the agents. Each action an agent $i$ can take is one of:

$$A^{i} = (\text{Noop, Move north, Move south, Move west, Move east, Pick-up})
$$

From the actions above, "Noop" action signifies the agent remaining idle, while "Pick-up" action allows the agent to collect food from an adjacent cell. The remaining actions result in the agent's moving in the corresponding direction on the 2D grid.

\subsubsection{Rewards}\hfill\\
An agent is rewarded only if they have collected at least one food item. 
When a food item is picked up, all agents adjacent to the food cell are rewarded accordingly, thus underlying the importance of cooperation in the game.

The reward given to each depends on the level of each participant agent so that higher-level agents get higher rewards for the same food picked up. The reward $i$ for each participant agent $i$ is calculated by the level of the food item picked up weighted by the level of the agent. This is then normalized by a factor of the sum of the other adjacent agent levels multiplied by the total number of foods that spawned in the current game.
All rewards are normalized so that all individual agent's rewards sum to one.

\subsection{Overcooked Environment Details}

In our experiments, we evaluated the performance of the proposed algorithms using the Overcooked environment, as described in \cite{overcooked}. 
The environment is a two-agents cooperative game, where the goal is to prepare and deliver various types of soups as efficiently as possible. The game takes place in a grid-based world, where each cell represents a distinct entity. The challenge is the ability of the agents to coordinate effectively, manage resources and optimize task execution. The agents' actions include navigating the grid, collecting ingredients, preparing the soups, and finally delivering them to a designated service area. The recipes available in the game involve two different ingredients, which can be combined in varying orders depending on the specific soup being prepared.

\subsubsection{Observations}\hfill\\
In our experiments, we utilized a variety of environmental layouts to test different cooperation challenges.
Specifically, we evaluated agent performance across three distinct layouts: "Cramped room", "Asymmetric advantages", and "Coordination ring". Each room focuses on different cooperation challenges. 

For the "\textbf{Cramped room}", the agents are positioned within a a single shared space. However, the grid layout is highly restrictive, so the environment tests the agents' ability to coordinate in confined spaces. 

For the "\textbf{Asymmetric advantages}", the layout is divided into separate rooms with each agent positioned in each one. This configuration tests the agents' high-level strategy decision-making ability. Successful coordination in this scenario relies on each agent’s capacity to optimize its actions based on the position and current actions of its co-players, highlighting the importance of inter-agent awareness and strategic planning.

In the "\textbf{Coordination ring}", similar to the cramped room layout, both agents are placed in a shared space but in a more spacious room. However, the pot (element needed for main interaction) is positioned at the top right corner of the layout, while all other interactions are positioned in the left bottom corner. The agents are tested whether they can coordinate and navigate the environment, ensuring they can execute distant but interconnected actions in a timely and efficient manner.

\begin{figure}[h]
    \centering
        \includegraphics[width=0.5\textwidth]{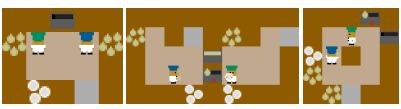}
    \caption{Sample snapshot of the Overcooked layouts. On the left is, "Cramped room", followed by the "Asymmetric advantages" and on the right is the "Coordination ring" layout.}
    \label{fig:overcooked}
\end{figure}

The three layouts can be seen in \autoref{fig:overcooked}, referenced by \cite{overcooked}.

In each timestep, both agents have full knowledge of the state of the environment. Each agent's observation space contains their position as well as the encoding of the next state. An overview of the observation matrix of an agent $i$ is : 
\[
O^{i} = {\text{(player\_i\_features, other\_player\_features, player\_i\_dist\_to\_other\_players, player\_i\_position)}}
\]
As observed from the above, their observation matrix contains both their own and their co-players' features, which is mostly their positional information to all objects in their environment. Specifically, according to the source code \cite{overcooked_github}, each player's features is a flattened list which contains:
    \begin{itemize}
    \item orientation: one-hot-encoding of the direction the player is currently facing. Possibilities are north, south, east, and west. This vector length is 4.
    \item obj: one-hot-encoding of the object currently being held. If nothing held, all 0s. Possibilities are onion, soup, dish, and tomato. This vector length is 4.
    \item wall\_{j}: 0, 1 boolean value of whether the player has a wall immediately in direction j. This vector length is 4.
    \item closest\_\{{onion|tomato|dish|soup|serving|empty\_counter}\}: (dx, dy) the distance to the item on both axes. (0, 0) if it is currently held. This vector length is $6*2$.
    \item closest\_soup\_n\_\{{onions|tomatoes}\}: an int indicating how many of the specified ingredients are inside the closest soup This vector length is 2.
    \item closest\_pot\_{j}\_exists: 0, 1, boolean value if there is a pot in the j direction. If not, then all the next pot features are 0. This vector length is 1.
    \item closest\_pot\_{j}\_\{{is\_empty|is\_full|is\_cooking|is\_ready}\}: {0, 1} boolean value for each pot statuses. O if there is no pot in j direction. This vector length is 4.
    \item closest\_pot\_{j}\_\{{num\_onions|num\_tomatoes}\}: int value indicating the number of each ingredient in jth closest pot. O if there is no pot in j direction. This vector length is 2.
    \item closest\_pot\_{j}\_cook\_time: int value indicating seconds remaining until soup is ready. -1 if there is no soup cooking. This vector length is 1.
    \item closest\_pot\_{j}: (dx, dy) distance on both axes to the pot in the j direction. (0, 0) of no pot on j direction. This vector length is 2.
    \end{itemize}

Note that the final number of closest\_pot features is multiplied by the number of total pots in the environment, which is 2. So the total 
\begin{equation*}
\begin{aligned}
\|O^{i}\| & = \text{player\_i\_features + other\_player\_features + player\_i\_relative\_position + player\_i\_absolute\_position} \\
& = \text{(2 * pot\_features + player\_i\_other\_features) + (2 * pot\_features + other\_player\_other\_features) }+ 2 + 2 \\
& = 2 * 20 + 56 = 96
\end{aligned}
\end{equation*}
An example of the total observation array of a sampled time step is given below:
\begin{lstlisting}[numbers=none]

array([ 0.,  1.,  0.,  0.,  0.,  0.,  0.,  0., -1., -1.,  0.,  0.,  0.,
        1.,  0.,  0.,  0.,  0.,  2.,  1.,  0.,  0.,  1.,  1.,  0.,  0.,
        0.,  0.,  0.,  0.,  1., -2.,  0.,  0.,  0.,  0.,  0.,  0.,  0.,
        0.,  0.,  0.,  0.,  1.,  0.,  1.,  1.,  0.,  0.,  0.,  0.,  0.,
        0.,  0.,  1.,  0.,  0.,  0., -2.,  2.,  0.,  0.,  0.,  0.,  0.,
        2.,  0.,  0.,  1.,  1.,  0.,  0.,  0.,  0.,  0.,  0., -1., -1.,
        0.,  0.,  0.,  0.,  0.,  0.,  0.,  0.,  0.,  0.,  1.,  0.,  1.,
        0.,  2., -1.,  1.,  2.], dtype=float32), 
\end{lstlisting}
This array contains the observation space, which is of size $(96)$, that belongs to a participating agents. All the values of the observation array also correspond to the fields explained earlier, which correspond to their features, the other player's features, their relative position, and lastly their absolute position on the grid world. For example, the first 4 values $\{0, 1, 0, 0\}$ represent agent's orientation in one-hot encoding. This means the agent is facing south. The next 4 numbers $\{0, 0, 0, 0\}$ correspond to the one-hot encoding of the object the agent holds, and since all are zero values, the agent does not hold any object. The rest of the values can be interpreted as explained in the individual fields of the player's features. Finally, the last 4 values correspond to the relative and absolute position. So $(2, -1)$ means that the 2nd agent is two positions on the x-axis further, but 1 position on the y-axis below. The last 2 values $(1, 2)$ show that the agent is in position $(1, 2)$ of the grid.

\subsubsection{Actions}\hfill\\
At each time step, the actions an agent can take correspond to the available directions they can move to. The agent takes 6 discrete actions as seen below:
\[
A^{i} = {(\text{Move north, Move south, Move east, Move west, Noop, Interact})}
\]

The last action is not related to navigation of the agent, but if selected it interacts with a nearby object.

\subsubsection{Rewards}\hfill\\
The reward types available are shaped or sparse and are jointly distributed to both agents. In our experiments, we have selected the sparse type, which is given to the agents only when a soup is delivered.
Once it is delivered, an agent is rewarded 0 points, if the soup is not made according to the available recipes for the current game. If the recipe of the soup is in the bonus orders, it receives a weighted reward of the base score value from the recipe multiplied by the additional bonus score of 2. Otherwise, the agent is rewarded with the calculated base score of 20. While the option for bonus orders is available, our selected layouts do not include any.
So the final reward for each agent is 0 if the soup was not created with the accepted number and order of ingredients, otherwise, the agent is rewarded 20 points.


\subsection{Pressureplate Environment Details}

\begin{figure}[h]
    \centering
        \includegraphics[width=0.2\textwidth]{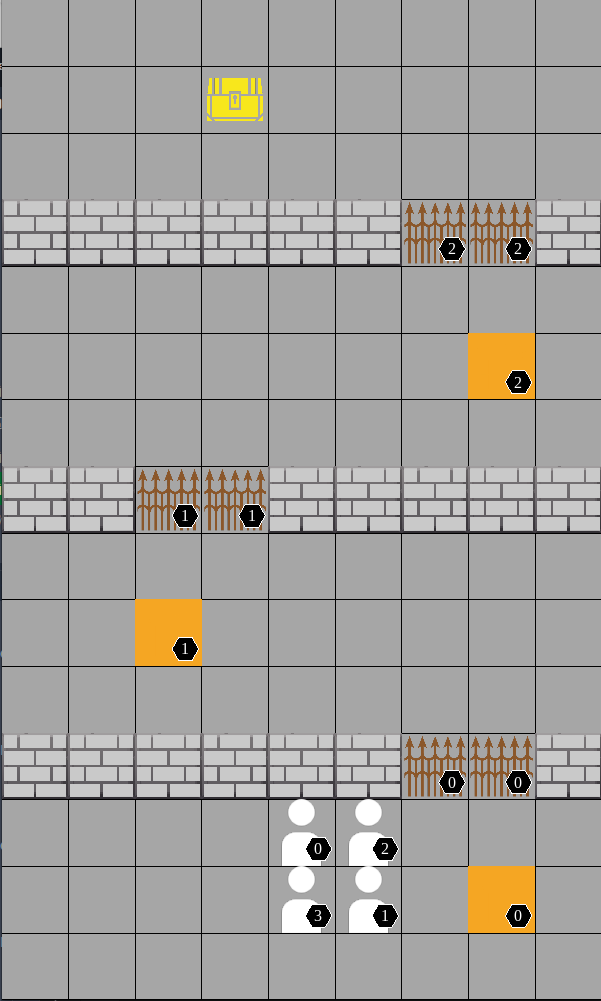}
    \caption{Snapshot of the Pressureplate environment. }
    \label{fig:pressureplate}
\end{figure}

In our experiments, we utilized the Pressureplate environment \cite{pressureplate_github}, which is a grid-world environment requiring agent cooperation. To progress and succeed, agents must collaborate by ensuring that the agent with the corresponding ID remains on a pressure plate in each room, which unlocks the door for the remaining agents to move to the next room. This process continues until all agents are on their respective plates, and the final agent reaches the treasure chest, which signifies the goal in the last room.  sample snapshot of the Pressureplate environment can be seen in \autoref{fig:pressureplate}.

\subsubsection{Observations}\hfill\\
In each new level, the grid map is divided into rooms by strategically placed walls, with each room containing a pressure plate and a corresponding locked door. At the start of each game, agents spawn in the southernmost room, with each assigned a unique ID that matches one of the plates.
Agents have partial observability of the environment, depending on its sensor range, specified by the parameter "sight". An example of an agent's observation array at a sampled time-step is given as:

\begin{lstlisting}[numbers=none]

array([ 0.,  0.,  0.,  0.,  0.,  0.,  1.,  1.,  0.,  0.,  0.,  1.,  1.,
        0.,  0.,  0.,  0.,  0.,  0.,  0.,  0.,  0.,  0.,  0.,  0.,  0.,
        0.,  0.,  0.,  0.,  0.,  0.,  0.,  0.,  0.,  0.,  0.,  0.,  0.,
        1.,  0.,  0.,  0.,  0.,  0.,  0.,  0.,  0.,  0.,  0.,  0.,  0.,
        0.,  1.,  1.,  0.,  0.,  0.,  0.,  0.,  0.,  0.,  0.,  0.,  0.,
        0.,  0.,  0.,  0.,  0.,  0.,  0.,  0.,  0.,  0.,  0.,  0.,  0.,
        0.,  0.,  0.,  0.,  0.,  0.,  0.,  0.,  0.,  0.,  0.,  0.,  0.,
        0.,  0.,  0.,  0.,  0.,  0.,  0.,  0.,  0.,  5., 13.], dtype=float32), 
\end{lstlisting}

The observation array of each agent contains a combination of 5 flattened observation sub-arrays of dimensions $sight \times sight$ each. Each observation sub-array represents the separate layout of a different element in the environment and the grid that is observable by the agent around them. These elements include the positions of other agents, walls, plates, doors, and goal (treasure chest), giving a total of 5 distinct observation sub-arrays per agent. 
If an element of the selected layout exists in one of these grid cells, then the value of that cell position on the respective sub-array will be 1, otherwise 0. 
After the layout sub-arrays are constructed, they are flattened to a final observation array of the agent. Finally, the agent's position coordinates $(x, y)$ are concatenated in the end, and thus form the final observation array. 

The total observation array's dimension is : $$\|O^{i}\| = N * (5 * (sight \times sight) + 2)$$
where N is the number of agents.

\subsubsection{Actions}\hfill\\
At each time step, each agent $i$ can select from 5 possible discrete actions, corresponding to their navigation on the grid-world map. These possible actions are:
\[
A^{i} = (\text{Move up, Move down, Move left, Move right, Noop})
\]

From the above, the last action represents idleness while the rest results in moving the agent by one cell in the chosen direction.

\subsubsection{Rewards}\hfill\\
Reward is independent of the actions of other participating agents.
At each time step, the distance between each agent's position and their corresponding pressure plate is calculated and their reward is then assigned to them, based on this distance. If the agent is in the same room as their respective plate, the reward will be the negative normalized Manhattan distance between the agent and the plate. However, if the agent is in a different room, the reward is calculated based on the number of rooms separating the agent from the room that contains their plate. Maximum rewards are given when all agents are in the correct room and the final agent has reached the treasure chest.

\subsection{Capture Target Environment Details}
\begin{figure}[h]
    \centering
        \includegraphics[width=0.3\textwidth]{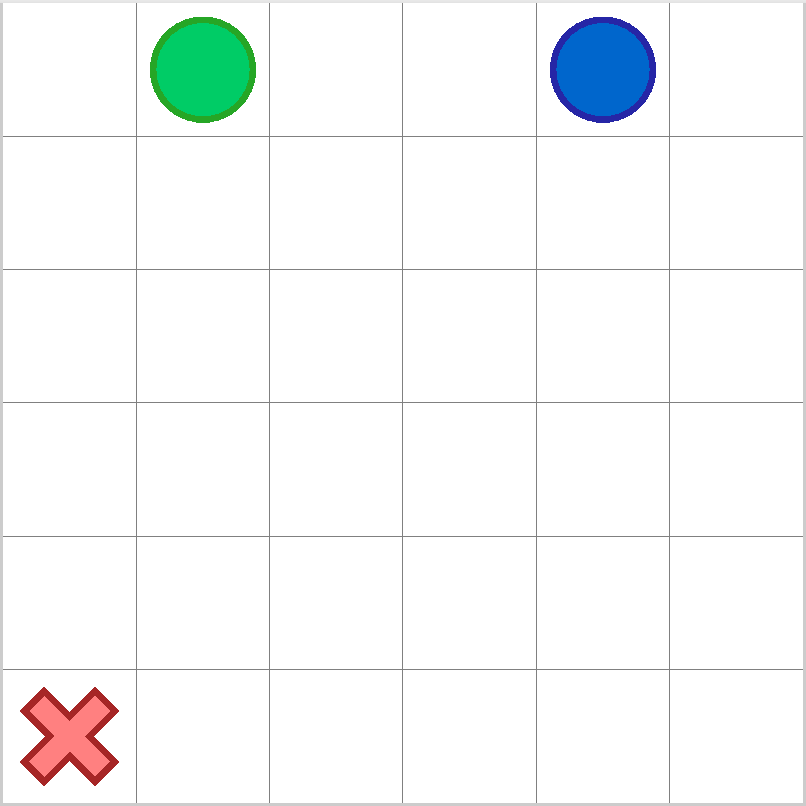}
    \caption{CaptureTarget environment. Two agents learn to capture the target simultaneously.}
    \label{fig:capturetarget}
\end{figure}

Agents in the Capture Target environment must learn the grid's transition dynamics to achieve simultaneous target capture, as quickly as possible. Both agents and targets move inside the grid. Agents observe the targets with a flickering probability at each time step. In the default Capture Target setup, which is presented in \autoref{fig:capturetarget}, there are 2 agents and 1 target, moving inside a 6X6 grid. The target's observation flickering probability is set to 0.3.

\subsubsection{Observations}\hfill\\
By default, the observations' array consists of the agents' and their targets' normalized positions on the grid. There is also an option for one-hot encoded observations. Targets' positions may differ from the actual ones, due to the flickering probability applied. In the 2 agents, 1 target scenario, the observation array per time step is of size $2 x 4$. For each agent $i$, the observation vector is 

$$O^{i}=(\text{x\_position}, \text{y\_position }, \text{target\_x\_position }, \text{target\_y\_position })$$
An example of an agent's observation array of a sample time step in a 6×6 grid is given below:
\begin{lstlisting}[numbers=none]
array([ 0.4  0.4  0.4  1.6], dtype=float32)
\end{lstlisting}
\subsubsection{Actions}\hfill\\
At each time step an agent can move in one of the available directions. An agent $i$ can pick one of the following actions: 
\[
A^{i} =(\text{Move north, Move south, Move west, Move east, No move})
\]


\subsubsection{Rewards}\hfill\\
In the successful scenario, where the agents capture the targets simultaneously, they are given +1 reward. At any other time step, the reward received is 0.

\subsection{Box Pushing Environment Details}
\begin{figure}[h]
    \centering
        \includegraphics[width=0.3\textwidth]{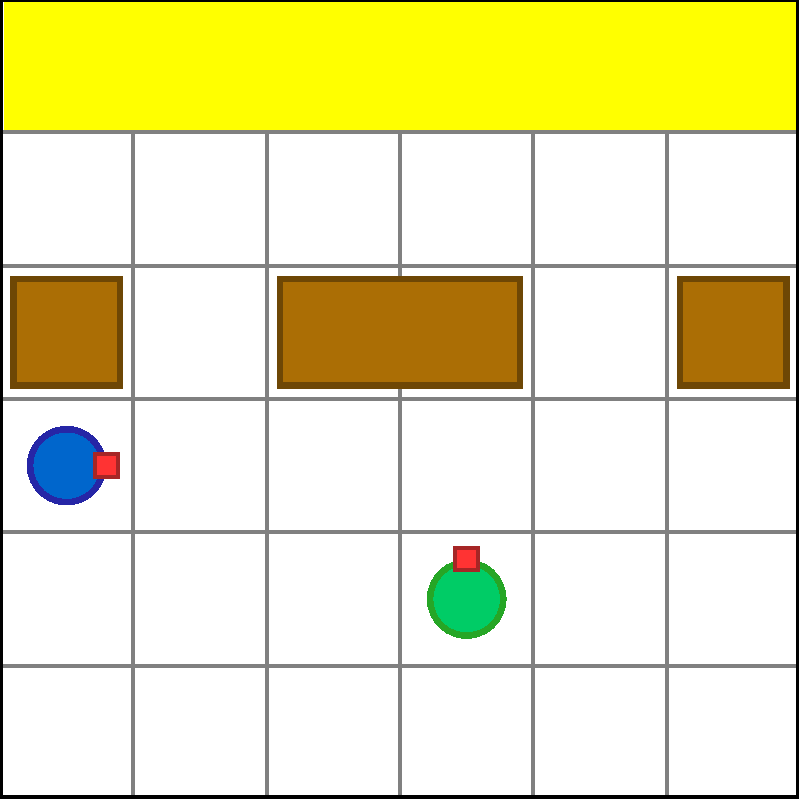}
    \caption{Snapshot of the Box Pushing environment.}
    \label{fig:boxpushing}
\end{figure}

The Box Pushing\cite{xiao_corl_2019} is a cooperative robotics problem introduced by \citeauthor{seuken2007} \cite{seuken2007} in which the main objective is two robots (agents) should cooperate and push a big box to the yellow space to get higher reward than pushing the other two small boxes that exist in the environment. In the default Box Pushing Target setup, which is depicted in \autoref{fig:boxpushing}, two agents, one big box and two small boxes exist and can move in a 6×6 grid.

\subsubsection{Observations}\hfill\\
The observation space of each agent consists of 5 elements in which the possible values are 0 and 1 indicating whether the element is in front of an agent (1) or not (0). An overview of the observation vector of an agent $i$ is: 

$$O^{i} = (\text{small\_box}, \text{large\_box}, \text{empty}, \text{wall}, \text{teammate})$$

An example of an agent's observation array of a sample time step in a 6×6 grid is given below:
\begin{lstlisting}[numbers=none]
array([0. 0. 1. 0. 0.], dtype=float32)
\end{lstlisting}
The values in this observation space show that the agent has an empty cell in front of them.
\subsubsection{Actions}\hfill\\
In every time step, an agent $i$ can choose one of the four available actions: 
\[
A^{i} = (\text{Move forward, Turn left, Turn right, Stay})
\]
The big box can be shifted when two robots are positioned facing it in adjacent cells and move forward simultaneously. The small boxes can be moved by one grid cell when a robot is positioned toward it and performs the move forward action.

\subsubsection{Rewards}\hfill\\
At every time step, the agents are rewarded by -0.1, and a successful push of the large box to the yellow space gives a +100 reward. In case a small box is pushed to the goal area, a +10 reward is returned to the agent. Unsuccessful scenarios that give a -5 penalty, are considered when an agent hits the boundary or when an agent alone tries to push the big box. 

\section{Additional Results}

\begin{figure}[H]
\centering

\setlength{\abovecaptionskip}{-10pt} 

\begin{table}[H]
\begin{tabular}{cccc}

\includegraphics[width=0.23\textwidth]{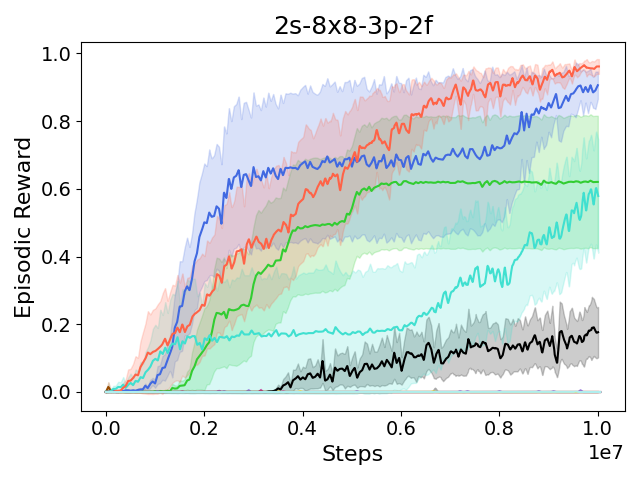} &
\includegraphics[width=0.23\textwidth]{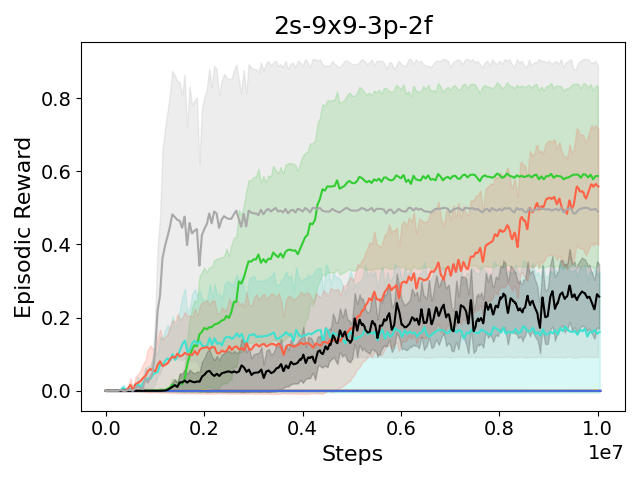} &
\includegraphics[width=0.23\textwidth]{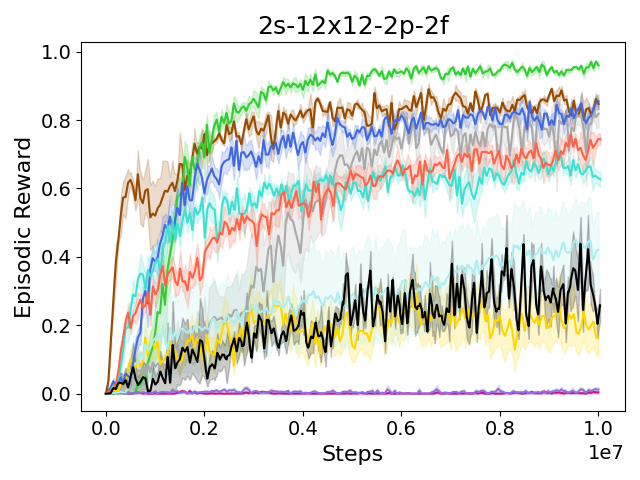} & \includegraphics[width=0.23\textwidth]{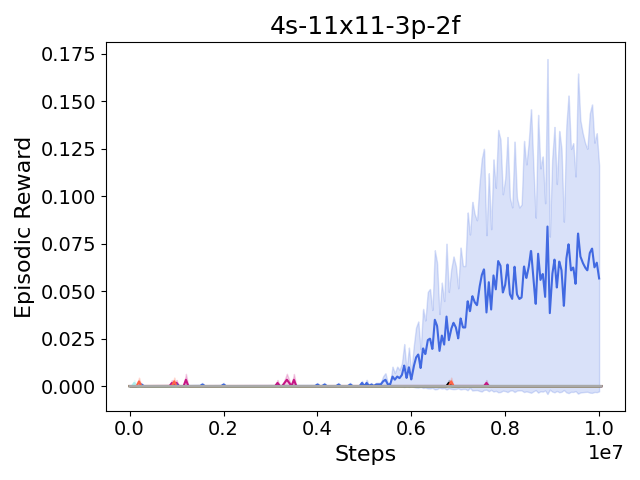} \\

\includegraphics[width=0.23\textwidth]{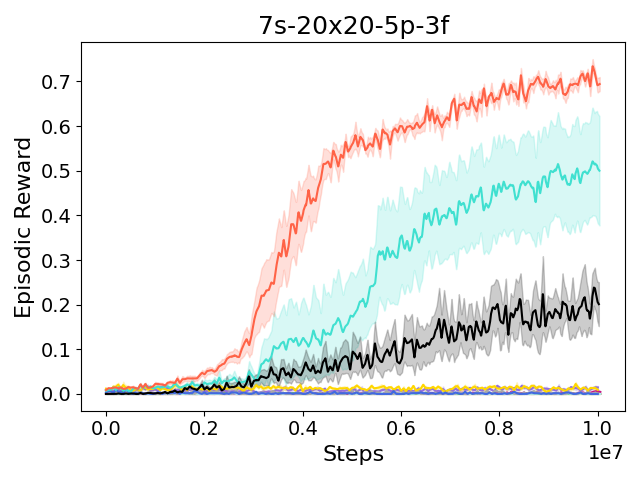} &
\includegraphics[width=0.23\textwidth]{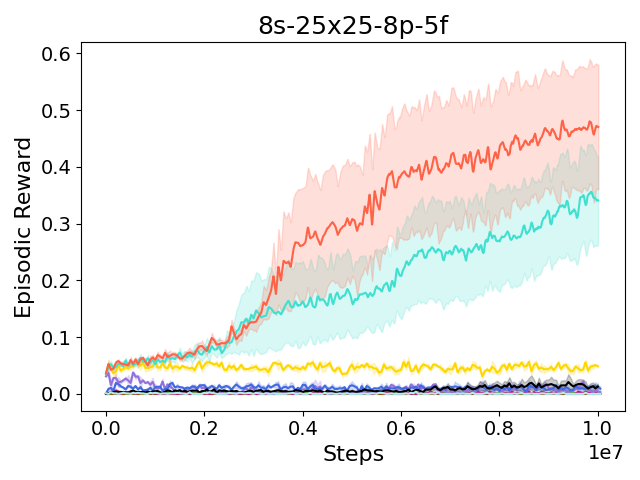} &
\includegraphics[width=0.23\textwidth]{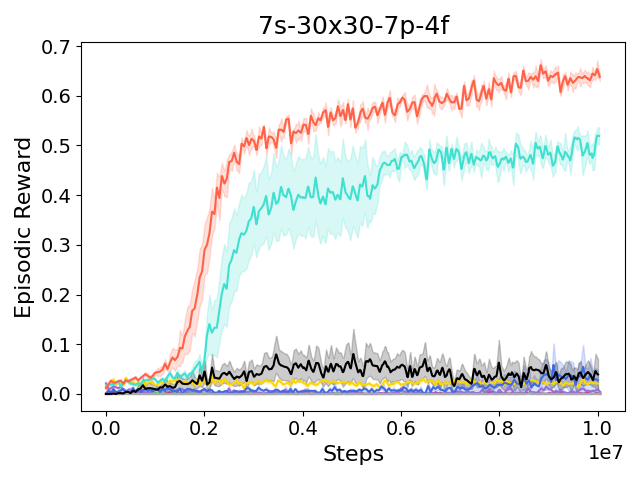} & \\

\multicolumn{4}{c}{\includegraphics[width=0.6\textwidth]{AAMAS-2025/curves/MARL_Legend.png}} \\

\end{tabular}
\end{table}
\caption{Episodic rewards of all 11 algorithms in LBF tasks rendering the
mean and the 75\% confidence interval over 5 different seeds.}
\label{fig:lbf_curves}
\end{figure}

\begin{figure}[H]
\centering

\setlength{\abovecaptionskip}{-10pt}

\begin{table}[H]
\begin{tabular}{ccc}

\includegraphics[width=0.23\textwidth]{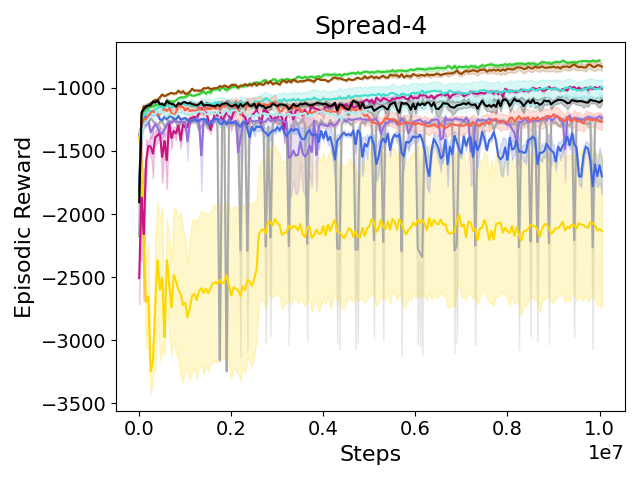} &
\includegraphics[width=0.23\textwidth]{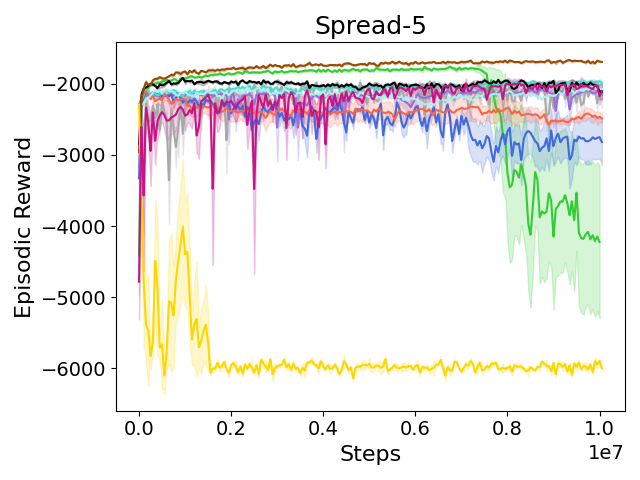} & \includegraphics[width=0.23\textwidth]{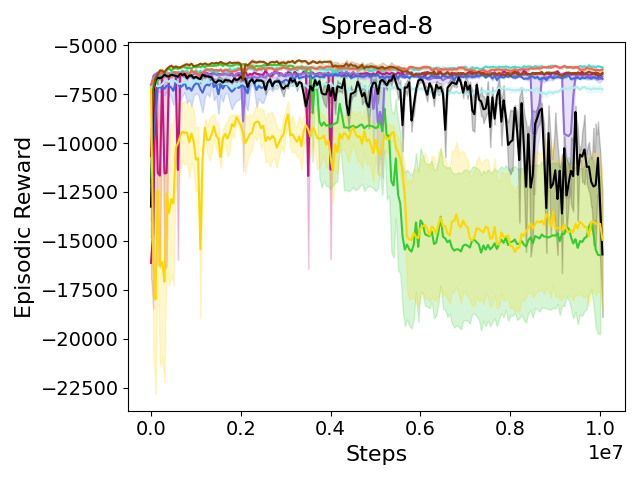} \\

\multicolumn{3}{c}{\includegraphics[width=0.6\textwidth]{AAMAS-2025/curves/MARL_Legend.png}} \\

\end{tabular}
\end{table}
\caption{Episodic rewards of all 11 algorithms in MPE tasks rendering the
mean and the 75\% confidence interval over 5 different seeds.}
\label{fig:mpe_curves}
\end{figure}

\begin{figure}[H]
\centering

\setlength{\abovecaptionskip}{-10pt} 

\begin{table}[H]
\begin{tabular}{ccc}

\includegraphics[width=0.23\textwidth]{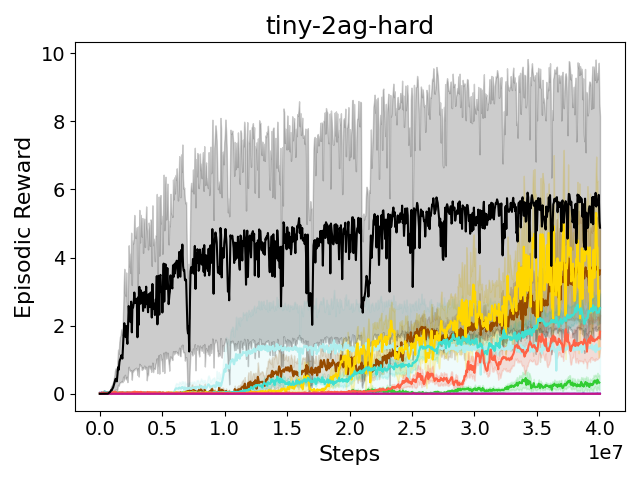} &
\includegraphics[width=0.23\textwidth]{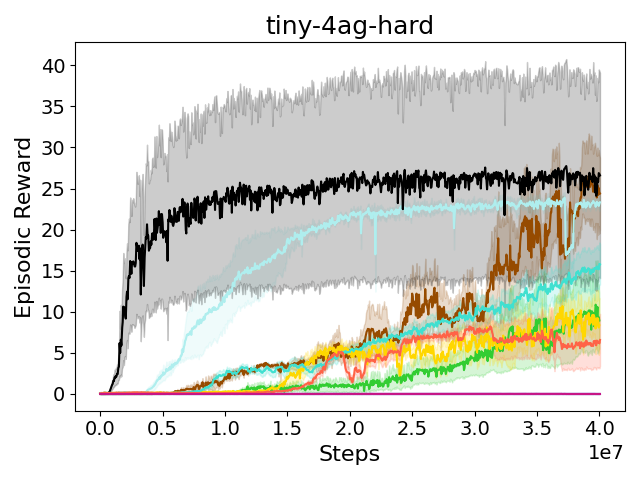} &
\includegraphics[width=0.23\textwidth]{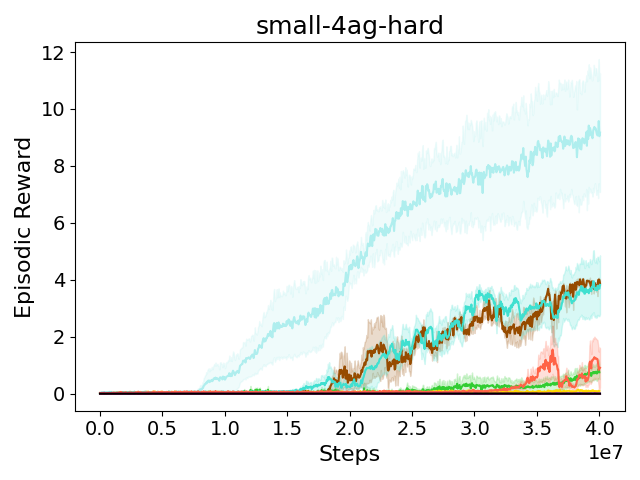} \\

\multicolumn{3}{c}{\includegraphics[width=0.6\textwidth]{AAMAS-2025/curves/MARL_Legend.png}} \\

\end{tabular}
\end{table}
\caption{Episodic rewards of all 11 algorithms in RWARE tasks rendering the
mean and the 75\% confidence interval over 5 different seeds.}
\label{fig:rware_curves}
\end{figure}

\begin{figure}[H]
\centering

\setlength{\abovecaptionskip}{-10pt} 

\begin{table}[H]
\begin{tabular}{ccc}

\includegraphics[width=0.23\textwidth]{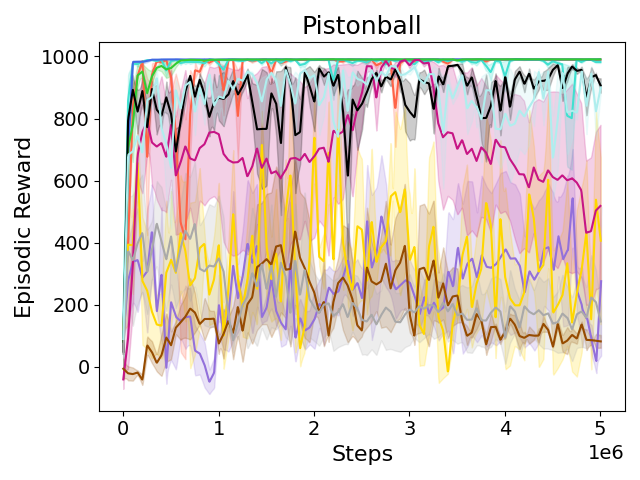} &
\includegraphics[width=0.23\textwidth]{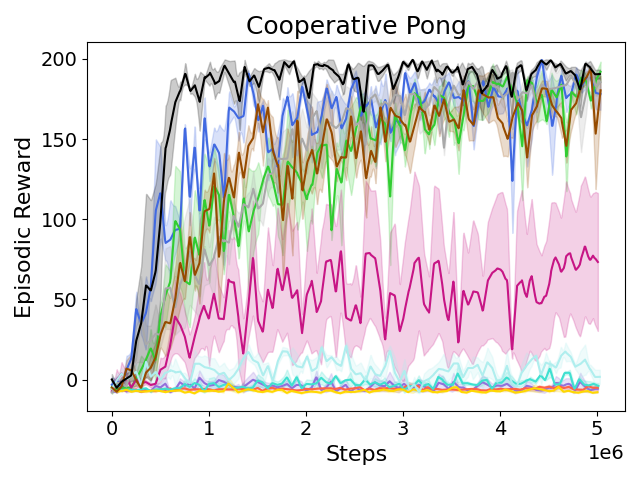} &
\includegraphics[width=0.23\textwidth]{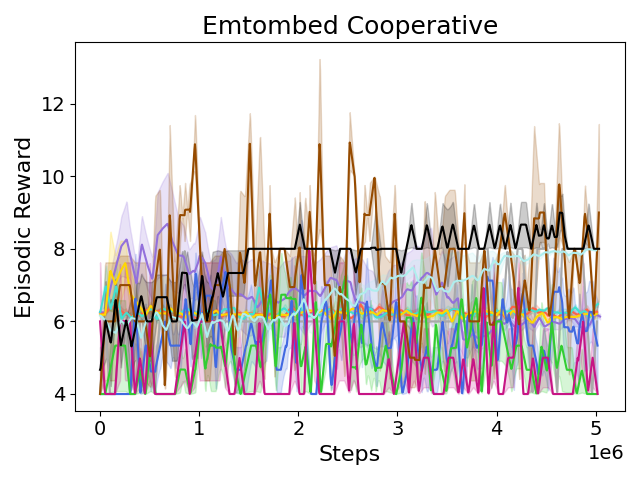} \\

\multicolumn{3}{c}{\includegraphics[width=0.6\textwidth]{AAMAS-2025/curves/MARL_Legend.png}} \\

\end{tabular}
\end{table}
\caption{Episodic rewards of all 11 algorithms in PettingZoo environments rendering the
mean and the 75\% confidence interval over 5 different seeds.}
\label{fig:pettingzoo_curves}
\end{figure}

\begin{figure}[H]
\centering

\setlength{\abovecaptionskip}{-10pt} 

\begin{table}[H]
\begin{tabular}{ccc}

\includegraphics[width=0.23\textwidth]{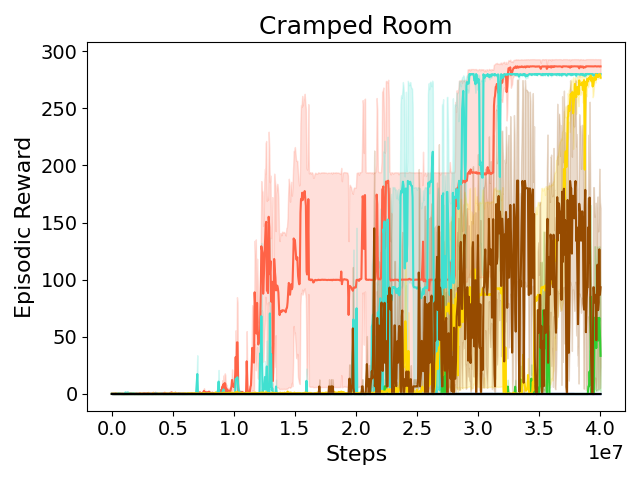} &
\includegraphics[width=0.23\textwidth]{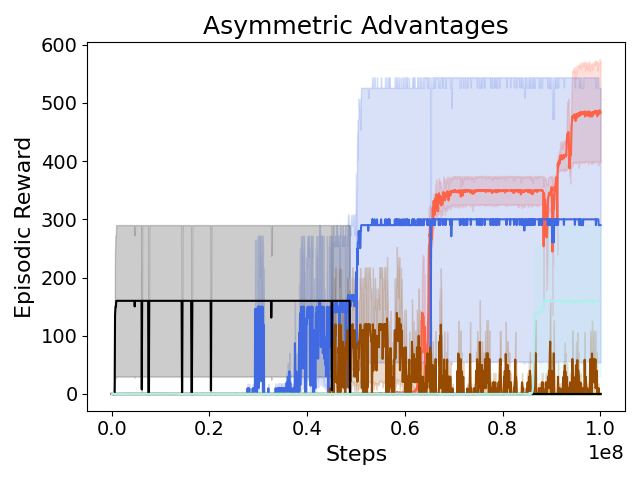} &
\includegraphics[width=0.23\textwidth]{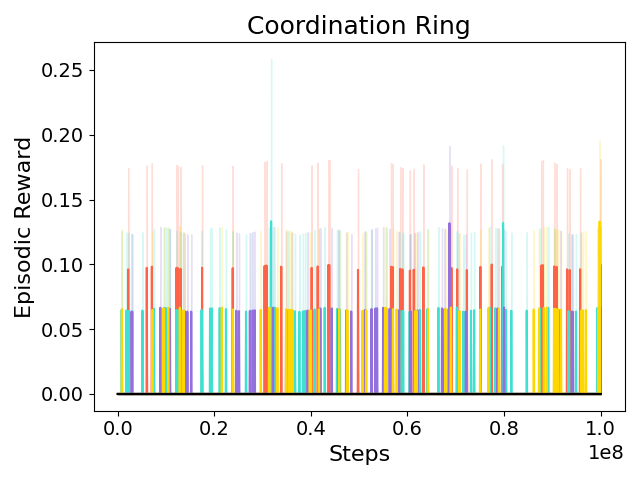} \\

\multicolumn{3}{c}{\includegraphics[width=0.6\textwidth]{AAMAS-2025/curves/MARL_Legend.png}} \\

\end{tabular}
\end{table}
\caption{Episodic rewards of all 11 algorithms in Overcooked environments rendering the
mean and the 75\% confidence interval over 5 different seeds.}
\label{fig:overcooked_curves}
\end{figure}

\begin{figure}[H]
\centering

\setlength{\abovecaptionskip}{-10pt} 

\begin{table}[H]
\begin{tabular}{cc}

\includegraphics[width=0.23\textwidth]{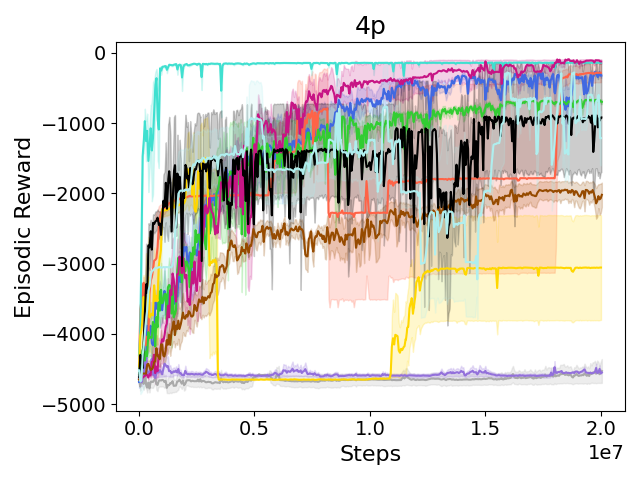} &
\includegraphics[width=0.23\textwidth]{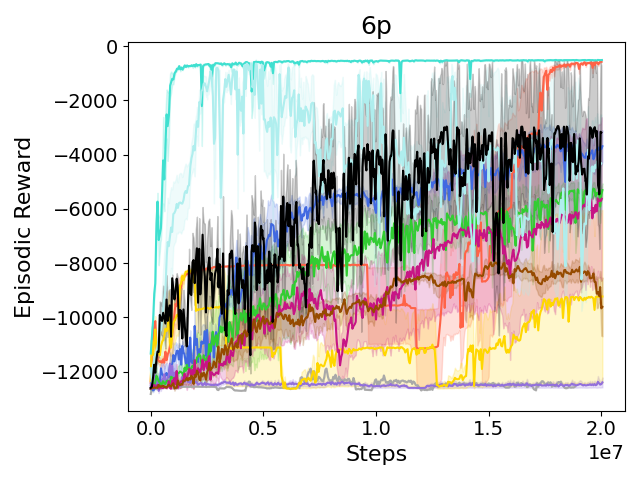} \\

\multicolumn{2}{c}{\includegraphics[width=0.6\textwidth]{AAMAS-2025/curves/MARL_Legend.png}} \\

\end{tabular}
\end{table}
\caption{Episodic rewards of all 11 algorithms in Pressure Plate tasks rendering the
mean and the 75\% confidence interval over 5 different seeds.}
\label{fig:pressureplate_curves}
\end{figure}

{
\small
\begin{table*}[h]
\centering

\begin{tabular}{@{} l *{6}{c} @{}}
\toprule
\textbf{Algorithms\textbackslash Environments} & \textbf{LBF} & \textbf{RWARE} & \textbf{Spread (MPE)} & \textbf{Petting Zoo} &\textbf{Overcooked} &  \textbf{Pressure Plate}\\
\midrule
\textbf{QMIX} & \(0.29 \pm 0.18\)  & \(0.00 \pm 0.00\)  & \(0.79 \pm 0.12\)  & \(0.40 \pm 0.28\)  & \(0.21 \pm 0.19\)  & \(0.86 \pm 0.11\) \\
\textbf{QPLEX} & \(0.32 \pm 0.17\)  & \(0.16 \pm 0.12\)  & \(0.64 \pm 0.30\)  & \(0.40 \pm 0.28\)  & \(0.06 \pm 0.06\)  & \(0.77 \pm 0.15\) \\
\textbf{MAA2C} & \(0.64 \pm 0.13\)  & \(0.13 \pm 0.06\)  & \(0.81 \pm 0.11\)  & \(0.34 \pm 0.31\)  & \(0.53 \pm 0.27\)  & \(0.97 \pm 0.01\) \\
\textbf{MAPPO} & \(0.46 \pm 0.11\)  & \(0.29 \pm 0.15\)  & \(0.83 \pm 0.11\)  & \(0.34 \pm 0.31\)  & \(0.19 \pm 0.18\)  & \(0.98 \pm 0.01\) \\
\textbf{HAPPO} & \(0.08 \pm 0.08\)  & \(0.42 \pm 0.22\)  & \(0.80 \pm 0.14\)  & \(0.34 \pm 0.30\)  & \(0.11 \pm 0.10\)  & \(0.97 \pm 0.01\) \\
\textbf{MAT-DEC} & \(0.22 \pm 0.08\)  & \(0.41 \pm 0.29\)  & \(0.80 \pm 0.14\)  & \(0.40 \pm 0.28\)  & \(0.00 \pm 0.00\)  & \(0.85 \pm 0.07\) \\
\textbf{COMA} & \(0.02 \pm 0.01\)  & \(0.00 \pm 0.00\)  & \(0.81 \pm 0.12\)  & \(0.23 \pm 0.21\)  & \(0.00 \pm 0.00\)  & \(0.32 \pm 0.26\) \\
\textbf{EOI} & \(0.07 \pm 0.05\)  & \(0.23 \pm 0.12\)  & \(0.50 \pm 0.25\)  & \(0.32 \pm 0.30\)  & \(0.19 \pm 0.18\)  & \(0.51 \pm 0.20\) \\
\textbf{MASER} & \(0.01 \pm 0.00\)  & \(0.00 \pm 0.00\)  & \(0.82 \pm 0.12\)  & \(0.37 \pm 0.30\)  & \(0.00 \pm 0.00\)  & \(0.79 \pm 0.17\) \\
\textbf{EMC} & \(0.23 \pm 0.16\)  & \(0.02 \pm 0.02\)  & \(0.81 \pm 0.12\)  & \(0.16 \pm 0.07\)  & \(0.00 \pm 0.00\)  & \(0.32 \pm 0.26\) \\
\textbf{CDS} & \(0.13 \pm 0.14\)  & \(0.42 \pm 0.26\)  & \(0.83 \pm 0.13\)  & \(0.21 \pm 0.11\)  & \(0.18 \pm 0.10\)  & \(0.60 \pm 0.20\) \\

\bottomrule
\end{tabular}
\caption{Mean average episodic rewards and 75\% confidence interval over each benchmark tasks for each algorithm.}
\label{tab:env-alg-summary}

\end{table*}
}

{
\small
\begin{table*}[t]
\centering
\begin{tabular}{@{} lccccccccccc @{}}

\toprule

\textbf{Tasks\textbackslash Algorithms} & \textbf{QMIX} & \textbf{QPLEX} & \textbf{MAA2C} & \textbf{MAPPO} & \textbf{HAPPO} & \textbf{MAT-DEC} & \textbf{COMA} & \textbf{EOI} & \textbf{MASER} & \textbf{EMC} & \textbf{CDS} \\

\midrule

\textbf{2s-8x8-3p-2f} & \(0d:7h\) & \(0d:11h\) & \(0d:1h\) & \(0d:2h\) & \(0d:4h\) & \(0d:3h\) & \(0d:1h\) & \(0d:6h\) & \(0d:14h\) & \(1d:8h\) & \(0d:13h\) \\

\textbf{2s-9x9-3p-2f} & \(0d:7h\) & \(0d:11h\) & \(0d:1h\) & \(0d:2h\) & \(0d:4h\) & \(0d:3h\) & \(0d:1h\) & \(0d:6h\) & \(0d:14h\) & \(2d:2h\) & \(0d:14h\) \\

\textbf{2s-12x12-2p-2f} & \(0d:6h\) & \(0d:11h\) & \(0d:1h\) & \(0d:2h\) & \(0d:3h\) & \(0d:3h\) & \(0d:1h\) & \(0d:5h\) & \(0d:11h\) & \(1d:6h\) & \(0d:10h\) \\

\textbf{4s-11x11-3p-2f} & \(0d:7h\) & \(0d:12h\) & \(0d:1h\) & \(0d:2h\) & \(0d:4h\) & \(0d:3h\) & \(0d:1h\) & \(0d:6h\) & \(0d:14h\) & \(2d:0h\) & \(0d:13h\) \\

\textbf{7s-20x20-5p-3f} & \(0d:9h\) & \(0d:14h\) & \(0d:1h\) & \(0d:2h\) & \(0d:6h\) & \(0d:4h\) & \(0d:1h\) & \(0d:6h\) & \(0d:20h\) & \(1d:12h\) & \(0d:20h\) \\

\textbf{8s-25x25-8p-5f} & \(0d:12h\) & \(0d:20h\) & \(0d:2h\) & \(0d:3h\) & \(0d:10h\) & \(0d:6h\) & \(0d:1h\) & \(0d:8h\) & \(1d:5h\) & \(4d:12h\) & \(1d:7h\) \\

\textbf{7s-30x30-7p-4f} & \(0d:11h\) & \(0d:18h\) & \(0d:2h\) & \(0d:3h\) & \(0d:9h\) & \(0d:6h\) & \(0d:1h\) & \(0d:7h\) & \(1d:2h\) & \(2d:18h\) & \(1d:3h\) \\

\textbf{Spread 4} & \(0d:10h\) & \(0d:15h\) & \(0d:2h\) & \(0d:2h\) & \(0d:6h\) & \(0d:4h\) & \(0d:2h\) & \(0d:7h\) & \(1d:1h\) & \(3d:20h\) & \(0d:19h\) \\

\textbf{Spread 5} & \(0d:12h\) & \(0d:18h\) & \(0d:2h\) & \(0d:3h\) & \(0d:8h\) & \(0d:5h\) & \(0d:2h\) & \(0d:9h\) & \(1d:6h\) & \(4d:5h\) & \(1d:0h\) \\

\textbf{Spread 8} & \(0d:23h\) & \(1d:8h\) & \(0d:4h\) & \(0d:5h\) & \(0d:14h\) & \(0d:9h\) & \(0d:4h\) & \(0d:18h\) & \(2d:5h\) & \(5d:3h\) & \(1d:19h\) \\

\textbf{tiny-2ag-hard} & \(1d:14h\) & \(2d:3h\) & \(0d:8h\) & \(0d:11h\) & \(0d:17h\) & \(0d:15h\) & \(0d:8h\) & \(1d:12h\) & \(1d:19h\) & \(18d:11h\) & \(2d:5h\) \\

\textbf{tiny-4ag-hard} & \(2d:2h\) & \(3d:1h\) & \(0d:10h\) & \(0d:13h\) & \(1d:6h\) & \(0d:23h\) & \(0d:10h\) & \(1d:22h\) & \(2d:12h\) & \(19d:3h\) & \(3d:13h\) \\

\textbf{small-4ag-hard} & \(2d:2h\) & \(3d:1h\) & \(0d:10h\) & \(0d:14h\) & \(1d:6h\) & \(0d:22h\) & \(0d:11h\) & \(1d:21h\) & \(2d:12h\) & \(19d:15h\) & \(3d:13h\) \\

\textbf{Pistonball} & \(2d:7h\) & \(4d:4h\) & \(4d:15h\) & \(4d:13h\) & \(32d:6h\) & \(3d:4h\) & \(4d:5h\) & \(1d:9h\) & \(3d:4h\) & \(6d:15h\) & \(4d:9h\) \\

\textbf{Cooperative Pong} & \(1d:15h\) & \(3d:17h\) & \(2d:20h\) & \(1d:13h\) & \(5d:15h\) & \(1d:0h\) & \(2d:5h\) & \(0d:14h\) & \(1d:5h\) & \(2d:20h\) & \(2d:0h\) \\

\textbf{Entomped Cooperative} & \(1d:4h\) & \(2d:14h\) & \(4d:10h\) & \(4d:5h\) & \(4d:8h\) & \(3d:6h\) & \(4d:1h\) & \(0d:23h\) & \(1d:18h\) & \(2d:12h\) & \(1d:11h\) \\

\textbf{Cramped Room} & \(1d:18h\) & \(2d:11h\) & \(0d:10h\) & \(0d:14h\) & \(0d:21h\) & \(0d:18h\) & \(0d:10h\) & \(1d:15h\) & \(2d:0h\) & \(13d:14h\) & \(2d:12h\) \\

\textbf{Asymmetric Advantages} & \(4d:13h\) & \(6d:16h\) & \(1d:1h\) & \(1d:12h\) & \(2d:6h\) & \(2d:0h\) & \(1d:3h\) & \(4d:7h\) & \(5d:5h\) & \(-\) & \(6d:11h\) \\

\textbf{Coordination Ring} & \(4d:13h\) & \(6d:6h\) & \(1d:0h\) & \(1d:11h\) & \(2d:3h\) & \(1d:22h\) & \(1d:2h\) & \(4d:7h\) & \(5d:22h\) & \(-\) & \(6d:7h\) \\

\textbf{4p} & \(0d:20h\) & \(1d:10h\) & \(0d:4h\) & \(0d:6h\) & \(0d:14h\) & \(0d:10h\) & \(0d:4h\) & \(0d:17h\) & \(1d:2h\) & \(9d:1h\) & \(1d:16h\) \\

\textbf{6p} & \(1d:1h\) & \(1d:18h\) & \(0d:4h\) & \(0d:8h\) & \(0d:20h\) & \(0d:14h\) & \(0d:5h\) & \(0d:20h\) & \(1d:10h\) & \(9d:13h\) & \(2d:10h\) \\

\bottomrule

\end{tabular}
\caption{Training times of all 11 algorithms in all benchmarks.}
\label{tab:training_times}
\end{table*}
}

\newpage

\section{Hyperparameters}
In this section, the hyperparameters used for each algorithm are presented in separate tables. We note that only in \textit{Spread-4} and \textit{Spread-5}, QMIX, MAA2C, COMA and MAPPO use non-sharing policy parameters.

\begin{table}[h]
    \centering
    \begin{minipage}{0.45\textwidth}
    \centering
        \caption{Hyperparameters for MAPPO}
        \begin{tabular}{cc}
            \toprule[1.5pt]
             Name & Value  \\
            \midrule
            agent runner & parallel(10)\\
            optimizer &  Adam\\
            batch size & 10\\
            hidden dimension & 64  \\
            learning rate & 0.0005  \\
            reward standardisation & True \\
            network type& GRU \\
            entropy coefficient & 0.01 \\
            target update & 200 \\
            buffer size & 10  \\
            $\gamma$ (discounted factor) & 0.99 \\
            observation agent id & True\\
            observation last action & True\\
            n-step& 5\\
            epochs & 4 \\
            clip & 0.2 \\
            \bottomrule[1.5pt]
        \end{tabular}
        
    \end{minipage}%
    \hspace{0.05\textwidth} 
    \begin{minipage}{0.45\textwidth}
        \centering
        \caption{Hyperparameters for MAA2C}
        \begin{tabular}{cc}
            \toprule[1.5pt]
             Name & Value  \\
            \midrule
            agent runner & parallel(10)\\
            optimizer &  Adam\\
            batch size & 10 \\
            hidden dimension & 64  \\
            learning rate & 0.0005  \\
            reward standardisation & True \\
            network type& GRU \\
           entropy coefficient & 0.01 \\
            target update & 200 \\
            buffer size & 10  \\
            $\gamma$ (discounted factor) & 0.99 \\
            observation agent id & True\\
            observation last action & True\\
            n-step& 5\\
            \bottomrule[1.5pt]
        \end{tabular}
    \end{minipage}
\end{table}

\begin{table}[h]
    \centering
    \begin{minipage}{0.45\textwidth}
        \centering
        \caption{Hyperparameters for QMIX}
        \begin{tabular}{cc}
            \toprule[1.5pt]
             Name & Value  \\
            \midrule
            agent runner & episode\\
            batch size & 32\\
            optimizer & Adam \\
            hidden dimension & 64  \\
            learning rate & 0.0005  \\
            reward standardisation & True \\
            network type& GRU \\
            evaluation epsilon & 0.0 \\
            epsilon anneal & 50000 \\
            epsilon start & 1.0 \\
            epsilon finish & 0.05 \\
            target update & 200 \\
            buffer size & 5000  \\
            $\gamma$ (discounted factor) & 0.99 \\
            observation agent id & True\\
            observation last action & True\\
            mixing network hidden dimension & 32 \\
            hypernetwork dimension& 64\\
            hypernetwork number of layers & 2 \\
            \bottomrule[1.5pt]
        \end{tabular}
    \end{minipage}%
    \hspace{0.05\textwidth} 
    \begin{minipage}{0.45\textwidth}
        \centering
        \caption{Hyperparameters for QPLEX}
        \begin{tabular}{cc}
            \toprule[1.5pt]
             Name & Value  \\
            \midrule
            agent runner & episode\\
            optimizer & RMSProp \\
            batch size & 32 \\
            hidden dimension & 64  \\
            learning rate & 0.0005  \\
            reward standardisation & True \\
            network type& GRU \\
             evaluation epsilon &0.0 \\
            epsilon anneal & 200000 \\
            epsilon start & 1.0 \\
            epsilon finish & 0.05 \\
            target update & 200 \\
            buffer size & 5000  \\
            $\gamma$ (discounted factor) & 0.99 \\
            observation agent id & True\\
            observation last action & True\\
            mixing network hidden dimension & 32 \\
            hypernetwork dimension& 64\\
            hypernetwork number of layers & 2 \\
            \bottomrule[1.5pt]
        \end{tabular}
    \end{minipage}
\end{table}

\begin{table}[h]
    \centering
    
\end{table}
\begin{table}[h]
    \centering
   
    \begin{minipage}{0.45\textwidth}
        \centering
        \caption{Hyperparameters for MAT-DEC}
        \begin{tabular}{cc}
            \toprule[1.5pt]
             Name & Value  \\
            \midrule
            agent runner & parallel(10)\\
            optimizer &  Adam\\
            batch size & 10 \\
            hidden dimension & 64  \\
            learning rate & 0.0005  \\
            reward standardisation & False \\
            value standardisation & True\\
            actor network type & MLP \\
            critic network type & Transformer \\
            evaluation epsilon & 0.0 \\
            epsilon anneal & 50000 \\
            epsilon start & 1.0 \\
            epsilon finish & 0.05 \\
            target update & 200 \\
            buffer size & 10  \\
            $\gamma$ (discounted factor) & 0.99 \\
            entropy coefficient & 0.01 \\
            value loss coefficient & 1\\
            observation agent id & True\\
            observation last action & False\\
            number of attention heads & 1 \\
            number of blocks & 1 \\ 
            epoch & 15\\
            clip & 0.2\\
            $\lambda_{GAE}$ & 0.95 \\
            use Huber loss & True \\
            delta coefficient of Huber loss & 10.0\\
            max norm of gradients & 10.0\\
            number of mini-batches  & 1\\
            \bottomrule[1.5pt]
        \end{tabular}
    \end{minipage}
    \hspace{0.05\textwidth} 
    \begin{minipage}{0.45\textwidth}
    \centering
        \caption{Hyperparameters for CDS}
        \begin{tabular}{cc}
            \toprule[1.5pt]
             Name & Value  \\
            \midrule
            agent runner & episode\\
            optimizer & Adam \\
            batch size & 5000 \\
            hidden dimension & 128  \\
            learning rate & 0.0005  \\
            reward standardisation & True \\
            network type& GRU \\
            evaluation epsilon & 0.0 \\
            epsilon anneal & 50000 \\
            epsilon start & 1.0 \\
            epsilon finish & 0.05 \\
            target update & 200 \\
            buffer size & 5000  \\
            $\gamma$ (discounted factor) & 0.99 \\
            observation agent id & False\\
            observation last action & True\\
            n-step & 10 \\
            entropy coefficient & 0.001\\
            number of attention heads & 4 \\
            attention regulation coefficient & 0.001\\
            mixing network hidden dimension & 32 \\
            hypernetwork dimension& 64\\
            hypernetwork number of layers & 2\\
            $\beta_{1}$ & 0.5\\
            $\beta_{2}$ & 1.0\\
            $\beta$ & 0.1\\
            $\alpha$ & 0.6\\
            $\lambda$ & 0.1\\
            \bottomrule[1.5pt]
        \end{tabular}
    \end{minipage}

\end{table}
\begin{table}[h]
 \begin{minipage}{0.45\textwidth}
        \centering
        \caption{Hyperparameters for COMA}
        \begin{tabular}{cc}
            \toprule[1.5pt]
             Name & Value  \\
            \midrule
            agent runner & parallel(10)\\
            optimizer & Adam \\
            batch size & 10\\
            hidden dimension & 64  \\
            learning rate & 0.0003  \\
            reward standardisation & True \\
            network type& GRU \\
            entropy coefficient & 0.001 \\
            target update & 200 \\
            buffer size & 10  \\
            $\gamma$ (discounted factor) & 0.99 \\
            observation agent id & True\\
            observation last action & True\\
            n-step& 10\\
            \bottomrule[1.5pt]
        \end{tabular}
    \end{minipage}%
        \hspace{0.05\textwidth}
    \begin{minipage}{0.45\textwidth}
        \centering
        \caption{Hyperparameters for EOI}
        \begin{tabular}{cc}
            \toprule[1.5pt]
             Name & Value  \\
            \midrule
            agent runner & episode\\
            optimizer & Adam \\
            batch size & 10 \\
            hidden dimension & 128  \\
            learning rate & 0.0005  \\
            reward standardisation & True \\
            network type& GRU \\
            target update & 200 \\
            buffer size & 10  \\
            $\gamma$ (discounted factor) & 0.99 \\
            observation agent id & True\\
            observation last action & True\\
            n-step & 5 \\
            entropy coefficient & 0.01 \\
            classifier ($\phi$) learning & 0.0001\\
            classifier ($\phi$) batch size & 256\\
            classifier  ($\phi$) $\beta_{2}$ & 0.1 \\
            \bottomrule[1.5pt]
        \end{tabular}
    \end{minipage}%

        \end{table}
\begin{table}[h]
    \centering
    \begin{minipage}{0.45\textwidth}
        \centering
        \caption{Hyperparameters for EMC}
        \begin{tabular}{cc}
            \toprule[1.5pt]
             Name & Value  \\
            \midrule
            agent runner & episode\\
            optimizer &  RMSProp\\
            batch size & 32 \\
            hidden dimension & 64  \\
            learning rate & 0.0005  \\
            reward standardisation & True \\
            network type& GRU \\
            evaluation epsilon& 0.0 \\
            epsilon anneal & 50000 \\
            epsilon start & 1.0 \\
            epsilon finish & 0.05 \\
            target update & 200 \\
            buffer size & 5000  \\
            $\gamma$ (discounted factor) & 0.99 \\
            observation agent id & True\\
            observation last action & True\\
            episodic memory capacity & 1000000\\
            episodic latent dimension & 4\\
            soft update weight & 0.005\\
            weighting term $\lambda$ of episodic loss & 0.1\\
            curiosity decay rate ($\eta_{t}$) & 0.9\\
            number of attention heads & 4 \\
            attention regulation coefficient & 0.001\\
             mixing network hidden dimension & 32 \\
            hypernetwork dimension& 64\\
            hypernetwork number of layers & 2\\
            \bottomrule[1.5pt]
        \end{tabular}
    \end{minipage}%
    \hspace{0.05\textwidth} 
    \begin{minipage}{0.45\textwidth}
        \centering
        \caption{Hyperparameters for MASER}
        \begin{tabular}{cc}
            \toprule[1.5pt]
             Name & Value  \\
            \midrule
            agent runner & episode\\
            optimizer & RMSProp \\
            batch size & 1 \\
            hidden dimension & 64  \\
            learning rate & 0.0005  \\
            reward standardisation & False \\
            network type& GRU \\
            evaluation epsilon & 0.0 \\
            epsilon anneal & 50000 \\
            epsilon start & 1.0 \\
            epsilon finish & 0.05 \\
            target update & 200 \\
            buffer size & 5000  \\
            $\gamma$ (discounted factor) & 0.99 \\
            observation agent id & True\\
            observation last action & True\\
            mixing network hidden dimension & 32 \\
            representation network dimension & 128\\
            $\alpha$ & 0.5 \\
            $\lambda$ & 0.03\\
            $\lambda_{I}$ & 0.0008 \\
            $\lambda_{E}$ & 0.00006 \\
            $\lambda_{D}$ & 0.00014\\
            \bottomrule[1.5pt]
        \end{tabular}
    \end{minipage}
\end{table}

\begin{table}[h]
\centering
        \caption{Hyperparameters for HAPPO}
         \begin{tabular}{cc}
        \toprule[1.5pt]
         Name & Value  \\
        \midrule
        agent runner & parallel(10)\\
        optimizer &  Adam\\
        batch size & 10 \\
        hidden dimension & 64  \\
        learning rate & 0.0005  \\
        reward standardisation & False \\
        value standardisation & True\\
        network type& GRU \\
        entropy coefficient & 0.01 \\
        buffer size & 10  \\
        $\gamma$ (discounted factor) & 0.99 \\
        observation agent id & False\\
        observation last action & False\\
         epochs & 5 \\
         n-step & 1 \\
          max norm of gradients  & 10 \\
         number of mini-batches  & 1\\
         $\lambda_{GAE}$ & 0.95 \\
         clip & 0.2 \\ 
         use huber loss & True \\
        delta coefficient of huber loss & 10.0\\
        \bottomrule[1.5pt]
        \end{tabular}

\end{table}

\end{document}